%% file: main.tex
\documentclass{article}

\usepackage[T1]{fontenc}
\usepackage{microtype}
\usepackage{graphicx}
\usepackage{subfigure}
\usepackage{booktabs}
\usepackage{placeins}

\usepackage{amsmath,amssymb,amsthm,mathtools}
\usepackage[hypertexnames=false]{hyperref}
\usepackage{xurl}

\usepackage[preprint]{mlsys2025}
\hypersetup{pdfauthor={Zikang Wen, Yuning Zhang, Dong Yuan}}

\mlsystitlerunning{MotionServe: Efficient Multi-Robot Serving}

\begin{document}

\twocolumn[
\mlsystitle{MotionServe: Efficient Multi-Robot Serving by Exploiting Robot Execution Slack}

\begin{mlsysauthorlist}
\mlsysauthor{Zikang Wen}{sydney}
\mlsysauthor{Yuning Zhang}{sydney}
\mlsysauthor{Dong Yuan}{sydney}
\end{mlsysauthorlist}

\mlsysaffiliation{sydney}{The University of Sydney, Sydney, Australia}
\mlsyscorrespondingauthor{Dong Yuan}{dong.yuan@sydney.edu.au}

\vskip 0.3in

\begin{abstract}
As robotic arms undertake diverse manipulation tasks, a deployment may need to serve many task-specific vision--language--action (VLA) policies concurrently. Dedicating a GPU to each robot is costly, motivating shared-GPU inference. Low-rank adaptation (LoRA) allows compatible policies to share a frozen backbone with lightweight policy-specific adapters, reducing model-weight memory and enabling cross-policy batching. However, robot inference must keep pace with physical execution. With action chunking, robots continue executing buffered actions while awaiting new predictions, and the remaining buffered time provides each robot's \emph{execution slack} for serving its next request. Once this slack is exhausted, the robot stalls until new actions arrive, interrupting task progress. Observing earlier preserves more slack but can lead to disruptive replanning and lower task success, whereas delaying dispatch to form larger batches can improve GPU efficiency but consumes slack and risks stalls. To balance these competing demands, we present MotionServe, a shared-GPU serving system that coordinates robot observation timing with server-side batch formation. MotionExec selects observation triggers based on motion structure in predicted action chunks, allowing inference to overlap with remaining buffered execution. SlackBatch combines the resulting per-robot execution budgets with estimated outstanding GPU work to determine when to wait, which requests to batch, and when to dispatch them. We evaluate X-VLA on LIBERO and SimplerEnv's WidowX tasks and OpenVLA-OFT on LIBERO, with up to 96 simulated robots sharing one NVIDIA H200 GPU. At 96 robots, MotionServe reduces mean stall per episode by 43.4\% and 32.6\% and improves task success by 8.98 and 2.30 percentage points, respectively, over each model's lowest-stall baseline. Backbone sharing also reduces weight memory for 96 policies by 97.6\% and 91.1\%.
\end{abstract}

]

\printAffiliationsAndNotice{}

\section{Introduction}
\label{sec:introduction}
Robotic arms are widely deployed across automotive, electronics, and general manufacturing to perform diverse manipulation tasks~\citep{Keshvarparast2024}. A single automotive spot-welding operation can involve 42 robots across 13 workstations~\citep{LOPES20171033}. Vision--language--action (VLA) models offer a common foundation for manipulation, mapping visual observations and language instructions to actions and supporting adaptation to new robot settings through fine-tuning~\citep{pmlr-v270-kim25c,KimM1-RSS-25}. A fleet performing different tasks may therefore require multiple fine-tuned variants of a common VLA. As the fleet grows, serving these variants with a dedicated GPU for each robot becomes increasingly costly, motivating centralized inference on a shared GPU server. On a shared server, however, hosting each variant as a separate model instance duplicates backbone weights and splits requests across instances that cannot be batched together~\citep{MLSYS2024_906419cd}. Because these variants originate from the same VLA, fine-tuning them with low-rank adaptation (LoRA) resolves both issues. Each variant then becomes a lightweight adapter over one shared frozen backbone, which reduces model-weight memory and allows requests for different policies to be batched together~\citep{hu2022lora,pmlr-v270-kim25c}.

A growing body of multi-LoRA serving work provides the machinery for this sharing. S-LoRA and Punica batch requests that target different adapters over a common language-model backbone~\citep{MLSYS2024_906419cd,MLSYS2024_054de805}, and dLoRA further improves efficiency through dynamic adapter merging and request placement~\citep{298762}. Batching nevertheless trades latency for throughput. Larger batches amortize computation, yet accumulating requests and executing larger batches both lengthen response latency~\citep{201468}. For robots, this latency carries a direct physical cost. A robot whose action buffer runs out before new actions arrive must pause, interrupting task progress~\citep{NEURIPS2025_300ccb21}. \textbf{Robot serving thus cannot trade latency for throughput freely}. The time remaining in each robot's buffer defines a time budget for serving its next request. This budget differs across robots and shrinks as execution proceeds. Existing multi-LoRA serving systems do not consider such execution-dependent budgets when deciding which requests to batch and when to dispatch them.

Unlike application-defined latency targets in conventional model serving, the time available for robot inference depends on ongoing physical execution and observation timing. With action chunking, each inference produces a sequence of future actions, allowing a robot to observe and request its next chunk before completing the current one while continuing to execute buffered actions~\citep{NEURIPS2025_300ccb21}. We term the remaining buffered execution time at request generation \emph{execution slack}. This slack defines the time available for communication, queueing, inference, and result delivery before the robot stalls. Observing earlier leaves more slack for serving, but replanning from an intermediate motion state can disrupt execution and reduce task success~\citep{NEURIPS2025_300ccb21,nie2026pace}. Observing later reduces the risk of premature replanning but leaves less time for batching. Efficient multi-robot VLA serving therefore requires coordinating two coupled decisions: when each robot observes and how the server batches requests within the resulting execution budgets.

To balance timely robot updates with efficient GPU batching, we propose MotionServe, a multi-LoRA VLA serving system that coordinates robot observation timing with batch formation through execution slack. Its \emph{Motion-Guided Pipelined Action Execution} (MotionExec) selects observation triggers from predicted motion structure, allowing robots to request new actions while they continue executing buffered actions. Its \emph{Slack-Driven Asynchronous Batching} (SlackBatch) combines each robot's remaining slack with estimated outstanding GPU work to decide how long to defer dispatch and which requests to include. Together, these mechanisms enable independent robot updates and cross-policy batching on a shared backbone. With 96 simulated robots sharing one NVIDIA H200 GPU, MotionServe reduces mean stall by 43.4\% on X-VLA and 32.6\% on OpenVLA-OFT and improves task success by 8.98 and 2.30 percentage points over each model's lowest-stall baseline.  MotionServe also achieves the highest throughput at every fleet size, and backbone sharing reduces weight memory for 96 policies by 91.1--97.6\%. Our contributions are:
\begin{itemize}
    \item We propose MotionServe, a framework that brings multi-LoRA serving to multi-robot VLA deployment while exploiting robot execution slack to coordinate observation timing with batch formation. It hosts each robot's policy as an adapter over one shared VLA backbone and batches action-chunk requests across adapters.

    \item We develop MotionExec, a motion-guided Pipelined action execution mechanism that selects observation triggers from predicted action chunks while overlapping inference with buffered execution.

    \item We develop SlackBatch, a slack-driven scheduler that uses each robot's execution slack and outstanding GPU work to adapt batch dispatch timing and request selection.
\end{itemize}

\section{Related Work and Motivations}
\label{sec:background}

\subsection{Multi-Robot VLA Serving}
\label{sec:bg-adapters}

Multi-robot deployments require distinct VLA policies adapted to different robot embodiments and manipulation tasks~\citep{pmlr-v270-kim25c,KimM1-RSS-25}. Deploying each policy as an independent model instance duplicates backbone weights and separates requests across instances, limiting cross-policy batching. Keeping fewer checkpoints resident reduces memory use but incurs loading overhead for nonresident policies. Low-rank adaptation (LoRA) instead allows compatible policies to share a frozen backbone through lightweight policy-specific updates~\citep{hu2022lora}. At an adapted layer with frozen weight $W_\theta$, policy $\ell$ uses $W_\theta+\gamma_\ell B_\ell A_\ell$, where $\gamma_\ell$ is a scaling factor and the adaptation rank is much smaller than the weight dimensions. For policies adapted from the same frozen backbone, a multi-LoRA runtime stores the backbone once and applies the corresponding adapter to each request within a batch, enabling cross-policy inference. Let $M_\theta$ denote the backbone's weight memory, $M_\ell$ the adapter and other policy-specific weight memory (including action heads where applicable), and $M_{{\rm full},\ell}$ the memory of a complete policy checkpoint. For $N$ resident policies, the model-weight footprints are
\begin{equation}
 M_{\rm independent}=\sum_{\ell=1}^{N}M_{{\rm full},\ell},
 \qquad
 M_{\rm shared}=M_\theta+\sum_{\ell=1}^{N}M_\ell.
 \label{eq:bg-memory}
\end{equation}

Multi-LoRA serving systems enable heterogeneous-adapter batching over a shared language-model backbone. S-LoRA and Punica batch requests using different adapters~\citep{MLSYS2024_906419cd,MLSYS2024_054de805}, while dLoRA further improves efficiency through dynamic adapter merging and request placement~\citep{298762}. These systems focus on autoregressive language-model serving rather than action-chunk inference constrained by robot execution. In robotics, CORAL trains task-specific LoRA experts on a frozen VLA backbone and provides inference-time expert routing~\citep{luo2026coralscalablemultitaskrobot}, but does not address cross-robot batch scheduling. Robion supports multi-robot, multi-model VLA serving through stage co-execution, model placement, and traffic control~\citep{adamopoulos2026efficientvisionlanguageactionmanagementserving}, but uses model-homogeneous batches within each stage. These approaches do not bring multi-LoRA serving to multi-robot VLA deployment, nor do they coordinate observation timing with batch formation through each robot's execution slack.

\noindent\textbf{Motivation 1.} Shared VLA serving should support independent robot update schedules while pooling requests across policies. This motivates an architecture in which robot-side execution determines request generation and a common multi-LoRA runtime combines the resulting requests into cross-policy batches.

\begin{figure}[t]
 \centering
 \includegraphics[width=\columnwidth]{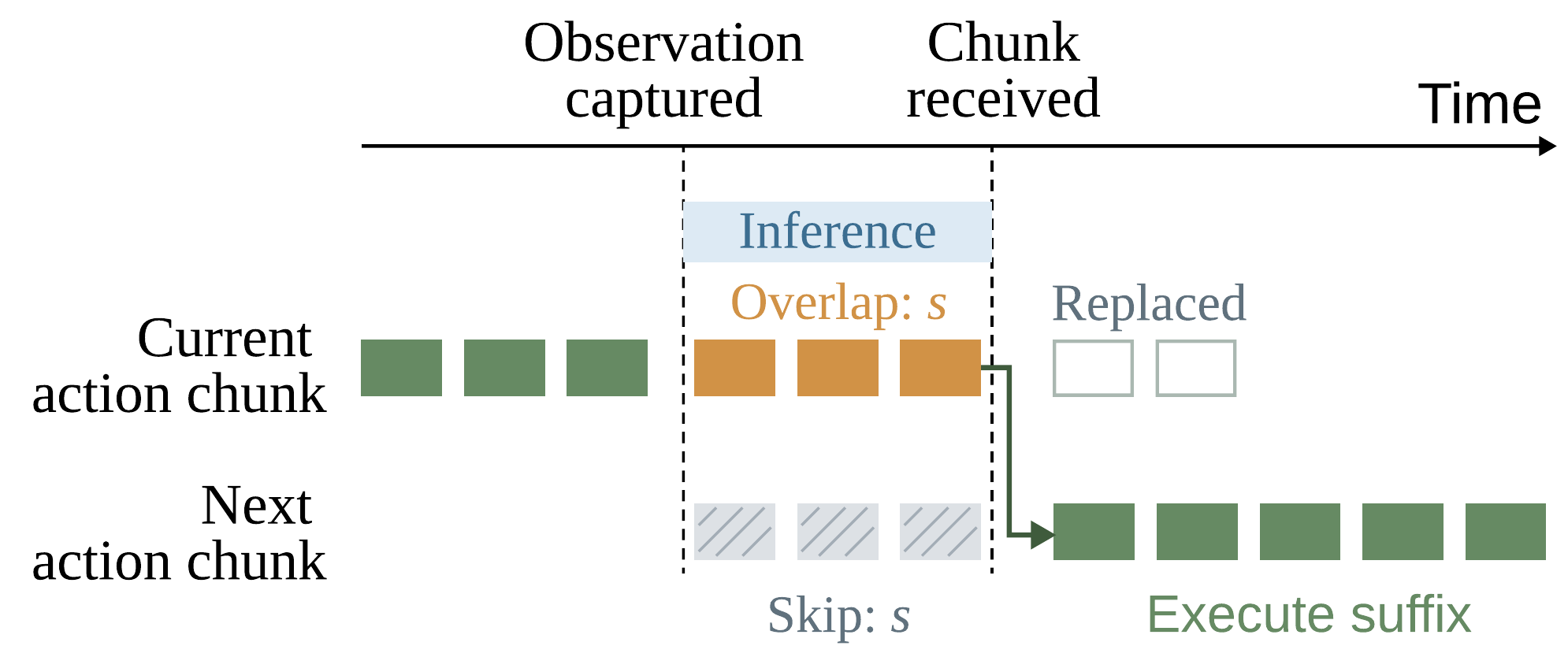}
 \caption{Pipelined action execution with overlap-and-skip handoff. Green blocks are executed actions. After observation, the robot executes $s$ buffered actions (orange) during inference, then skips the first $s$ actions of the returned chunk (gray) and executes its suffix. Each block is one action.}
 \label{fig:bg-async-overlap}
\end{figure}
\begin{table}[t]
\centering
\caption{Task success (\%) on LIBERO-10 for full-chunk observation and for Pipelined observation at a short prefix, long prefix, or motion-derived sweet spot. Bold marks the best result for each policy.}
\label{tab:bg-refresh-quality}
\vskip 0.15in
\begin{tabular}{lrrrr}
\toprule
Policy & Full & \shortstack{Short\\prefix} & \shortstack{Long\\prefix} & \shortstack{\textbf{Sweet}\\\textbf{spot}} \\
\midrule
TurboVLA & 80.8 & 63.0 & 80.4 & \textbf{82.2} \\
TBD-VLA & 96.6 & 78.2 & 93.0 & \textbf{98.2} \\
X-VLA & 80.4 & 51.4 & 80.6 & \textbf{85.4} \\
OpenVLA-OFT & 93.4 & 69.8 & 83.0 & \textbf{94.0} \\
ActionCodec-AR & 78.8 & 77.4 & 82.0 & \textbf{83.6} \\
\bottomrule
\end{tabular}
\end{table}
\begin{figure}[t]
 \centering
 \includegraphics[width=\columnwidth]{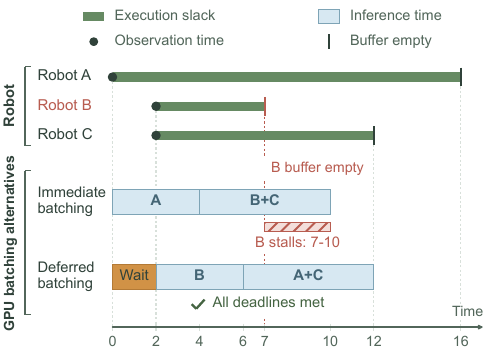}
 \caption{Execution slack guides batching. Robot rows show each robot's slack, and GPU rows compare two schedules for the same requests. Deferring A avoids B's stall and enables batching A with C.}
 \label{fig:bg-execution-slack}
\end{figure}
\begin{figure*}[t]
 \centering
 \includegraphics[width=0.95\textwidth]{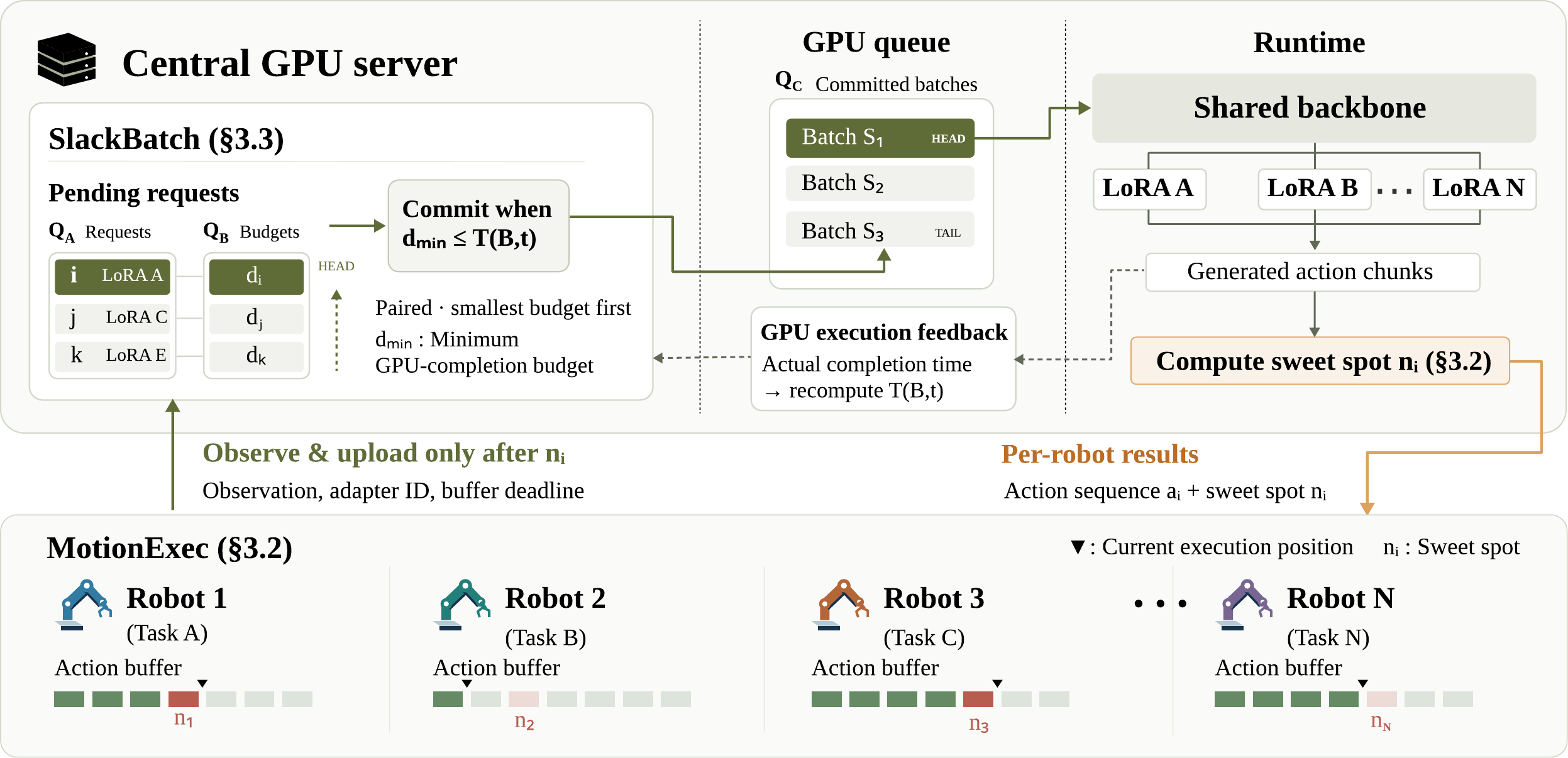}
 \caption{MotionServe overview. MotionExec (\S\ref{sec:sweet-spot-execution}) triggers requests at each robot's sweet spot, and SlackBatch (\S\ref{sec:slack-aware-batching}) batches them by comparing the minimum budget $d_{\min}$ with the estimated GPU time $T(B,t)$. Dashed arrows denote GPU completion feedback.}
 \label{fig:system-overview}
\end{figure*}

\subsection{Pipelined Action Execution and Observation Timing}
\label{sec:bg-refresh}

Action chunking allows a VLA policy to predict a sequence of future actions in a single inference request~\citep{KimM1-RSS-25}, so a robot can continue moving while it awaits the next prediction. \emph{Pipelined action execution} exploits this buffer by letting the robot capture a new observation and request its next chunk before exhausting the current one~\citep{NEURIPS2025_300ccb21}. As Figure~\ref{fig:bg-async-overlap} illustrates, the robot keeps executing buffered actions while the request is served, and the remaining buffered execution time forms its execution slack. When the result arrives, the robot skips the new chunk's first $s$ actions, where $s$ counts the policy actions completed since observation capture, and executes the remaining suffix. This handoff aligns the new chunk with completed execution and avoids idle waiting whenever the result arrives before the buffer is depleted. Because each robot follows its own execution progress, requests from different robots arrive independently at the shared server~\citep{bansal2026actionchunkschedulingbatched}, each carrying a different amount of slack.

Observation timing determines how much slack remains for serving. Earlier observation provides more slack but may trigger replanning before a coherent motion segment finishes, disrupting execution and reducing task success~\citep{NEURIPS2025_300ccb21,nie2026pace}. Table~\ref{tab:bg-refresh-quality} examines this tradeoff on LIBERO-10 across five VLA policies with different action-generation designs: TurboVLA~\citep{xie2026turbovlarealtimevisionlanguageactionmodel}, OpenVLA-OFT~\citep{KimM1-RSS-25}, X-VLA~\citep{ICLR2026_630d2938}, TBD-VLA~\citep{lee2026tbdvlatemporalblockdiffusion}, and ActionCodec-AR~\citep{dong2026actioncodecmakesgoodaction}. Full observes after completing the chunk, while Short prefix and Long prefix use fixed observation triggers. Sweet spot selects the earliest suitable observation point under a motion-derived progress criterion (Section~\ref{sec:sweet-spot-execution}). The latter three use the same overlap-and-skip protocol. Short prefix reduces success by 1.4--29.0 percentage points relative to Full, while Long prefix improves success over Short prefix for every policy but exceeds Full only for X-VLA and ActionCodec-AR. Sweet spot exceeds Full by 0.6--5.0 percentage points across all five policies, showing that well-placed early observation can preserve or improve task success.

Real-Time Chunking (RTC) overlaps execution with flow-policy inference using inpainting to improve continuity~\citep{NEURIPS2025_300ccb21}. Adaptive execution-horizon methods select replanning points using motion transitions (PACE)~\citep{nie2026pace}, action uncertainty (AAC)~\citep{Liang_2026_CVPR}, or self-attention (AutoHorizon)~\citep{10.1007/978-3-032-37196-6_33}. However, these methods do not coordinate observation timing with slack-aware batch formation across multiple robots.

\noindent\textbf{Motivation 2.} Observation timing should balance task progress, timely feedback, and the execution slack available for serving. This motivates selecting the earliest suitable observation point based on predicted motion structure, allowing robots to avoid premature replanning while retaining buffered actions for shared inference.

\subsection{Execution Slack for Asynchronous Batching}
\label{sec:bg-async}

Asynchronous robot execution produces independently arriving requests with different execution slack. Batching these requests can improve shared-GPU efficiency, but waiting to accumulate requests consumes the time available before robots run out of actions. Immediate dispatch avoids this delay, yet may occupy the GPU with less urgent requests and delay later arrivals whose buffers deplete sooner. Figure~\ref{fig:bg-execution-slack} illustrates this tradeoff. Robot A requests inference first while the GPU is idle, so first-come-first-served (FCFS), earliest-deadline-first (EDF), and any other work-conserving scheduler dispatch it immediately. When B arrives with less slack remaining, it must wait behind A and stalls. Deferring A instead allows the server to serve B first and then batch A with C, keeping all three robots moving in this example.

Deciding when to dispatch requires considering both the robots' execution slack and outstanding GPU work. After reserving time for result delivery, each request's execution slack leaves a remaining GPU-completion budget. For a candidate batch, we define its \emph{dispatch margin} as the smallest of these budgets minus the estimated preceding GPU work and inference time for that batch. A positive margin leaves room to accumulate more requests under the current latency estimates. As robots execute and GPU work progresses, this margin changes, requiring the server to adapt both dispatch timing and batch membership.

Clipper studies delayed batching with a configurable wait timeout~\citep{201468}, but this waiting mechanism does not adapt to robot execution slack. Armory uses lookahead over robot execution states to select batches under heterogeneous execution horizons~\citep{bansal2026actionchunkschedulingbatched}. However, it does not explicitly use execution slack to determine how long dispatch can be deferred to accumulate requests across policies.

\noindent\textbf{Motivation 3.} Batch formation should balance GPU efficiency against the risk of robot stalls. This motivates using remaining execution budgets and GPU progress to decide when to wait and which requests to serve together.

\section{System Design}
\label{sec:design}
\subsection{System Overview}
\label{sec:system-overview}
MotionServe couples MotionExec and SlackBatch through each robot's execution slack. As shown in Figure~\ref{fig:system-overview}, $N$ robots execute independently while a central server hosts a frozen VLA backbone, their LoRA adapters, and policy-specific heads. The two components interact through a compact interface: each request carries a buffer-depletion deadline, and each response returns the next sweet spot. MotionExec uses the motion-derived sweet-spot index $n_i$ to determine when robot $i$ requests new actions. The robot captures a fresh observation, uploads it with its adapter ID and deadline, and keeps executing buffered actions throughout communication, queueing, and inference. Since robots progress independently, their requests arrive asynchronously with different execution slack. Section~\ref{sec:sweet-spot-execution} details observation triggering and chunk handoff.

SlackBatch uses this slack to coordinate batch formation. It reserves each robot's offline-profiled result-delivery latency to obtain a GPU-completion budget, then orders pending requests ($Q_A$) and their budgets ($Q_B$) from most to least urgent. After accounting for outstanding GPU work and candidate batch inference time, SlackBatch waits while the full pending batch has a positive estimated dispatch margin. Otherwise, it commits to the FIFO queue $Q_C$ the largest prefix of deadline-ordered groups that can still finish within the most urgent budget, or the head group if no prefix can. The scheduler then reevaluates the remaining requests (Section~\ref{sec:slack-aware-batching}).

The runtime executes committed batches in FIFO order ($S_1$ to $S_3$ in Figure~\ref{fig:system-overview}) using multi-LoRA inference and feeds actual GPU completion times back to SlackBatch. For each robot $i$, the server computes a new sweet spot $n_i$ from the generated action chunk $A_i$ and returns both. MotionExec aligns the returned chunk with the actions completed since observation capture and replaces the robot's buffer with the remaining suffix, whose slack and sweet spot govern the next request. Result delivery and buffer replacement overlap with subsequent inference, allowing each robot's execution and inference cycle to advance independently.

\begin{figure}[t]
 \centering
\includegraphics[width=\columnwidth]{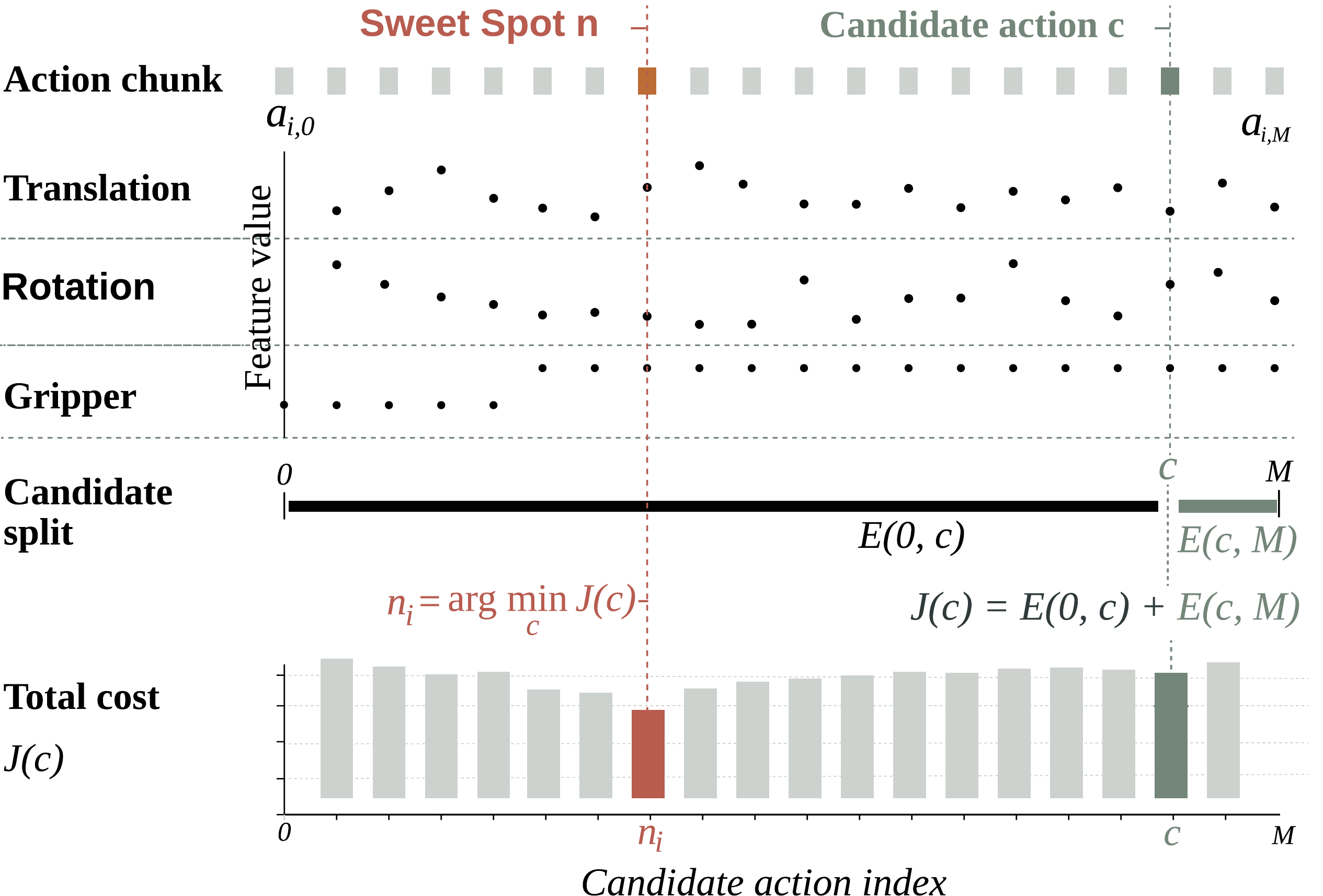}
 \caption{Sweet-spot selection in MotionExec. Each candidate split $c$ divides the chunk into two segments, and $J(c)$ sums their fitting costs over motion and gripper features. The minimum-cost index $n_i$ marks the action whose completion triggers observation.}
 \label{fig:sweet-spot-selection}
\end{figure}
\begin{figure}[t]
 \centering
 \includegraphics[width=\columnwidth]{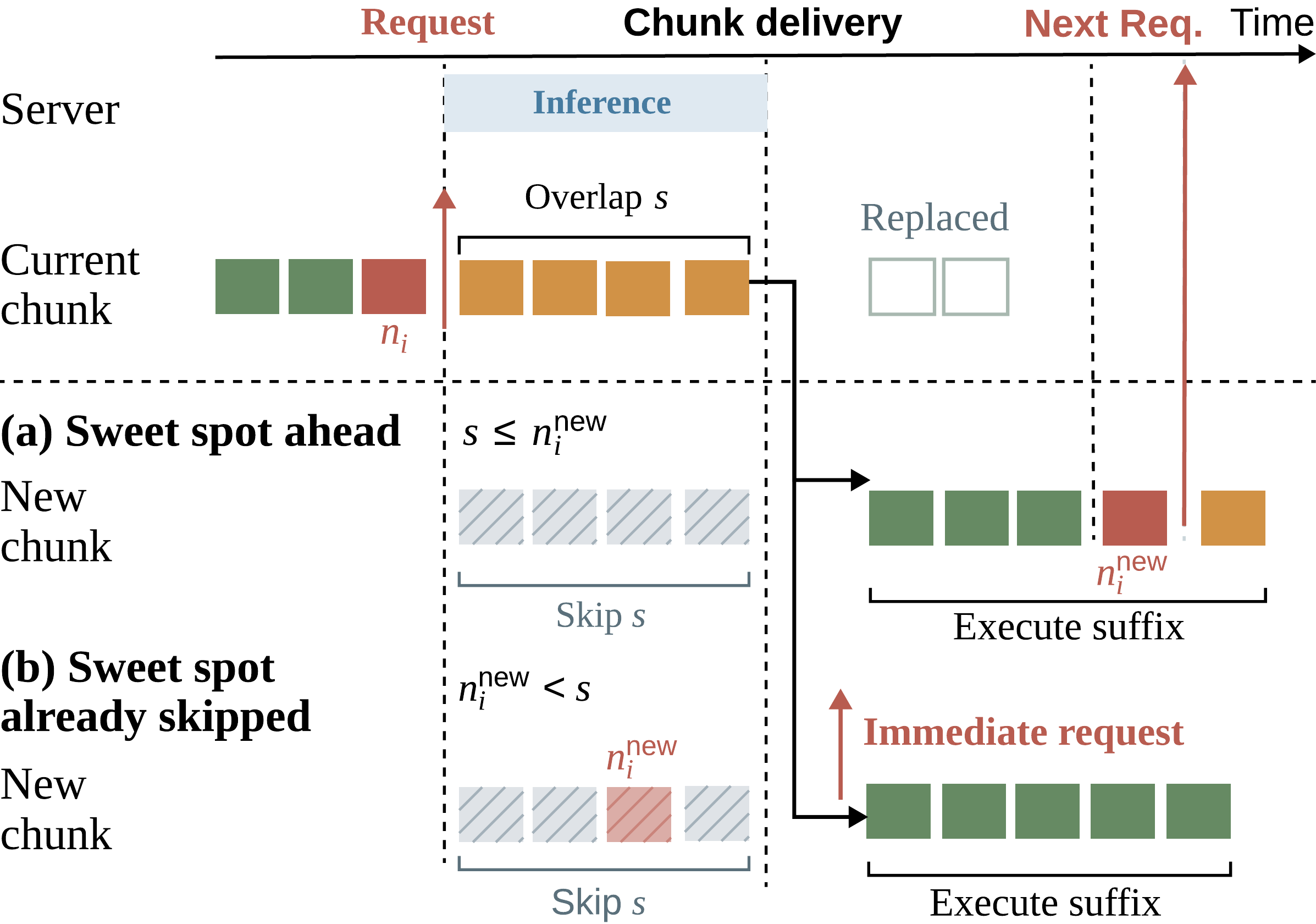}
 \caption{Pipelined action execution and chunk handoff in \textbf{MotionExec}. At handoff, the robot skips the returned chunk's first $s$ actions, where $s$ counts actions completed since observation. It requests again (a) after completing the new sweet-spot action, or (b) immediately after handoff if that action lies in the skipped prefix.}
 \label{fig:sweet-spot-execution}
\end{figure}
\subsection{Motion-Guided Pipelined Action Execution}
\label{sec:sweet-spot-execution}

MotionExec uses predicted motion to time each robot's next observation, then overlaps request processing with execution of the remaining actions. The action buffer thus links the observation trigger to the execution slack available for serving.

\paragraph{Choosing the observation trigger.}
For robot $i$, let $A_i=(a_{i,0},\ldots,a_{i,H_i-1})$ be a chunk of policy-fixed length $H_i$, where $a_{i,j}$ is the action at zero-based index $j$. MotionExec approximates a motion transition with a two-segment fit, as illustrated in Figure~\ref{fig:sweet-spot-selection}. Each chunk contains $M=H_i-1$ motion intervals. The six-dimensional feature $z_j$ concatenates normalized translation and rotation increments for interval $j$, and $g_j$ denotes the binary open/close gripper command at action $j$. Candidate action index $c$ separates intervals $[0,c)$ and $[c,M)$. For a segment $[l,r)$ with $0\le l<r\le M$, we define its minimum fitting cost as
\begin{align}
 E(l,r)={}&\min_{\alpha,\beta\in\mathbb R^6}
       \sum_{j=l}^{r-1}\|z_j-\alpha-\beta j\|_2^2 \nonumber\\
       &+\lambda_g\min_{\mu\in\mathbb R}\sum_{j\in G(l,r)}(g_j-\mu)^2.
 \label{eq:sweet-spot-error}
\end{align}
The first term fits an affine motion trend with intercept $\alpha$ and slope $\beta$, and the second fits a constant gripper value $\mu$, weighted by $\lambda_g=4$. The gripper-index set is $G(0,r)=\{0,\ldots,r\}$ for the first segment and $G(l,r)=\{l+1,\ldots,r\}$ for $l>0$, so each command is assigned once. Following cost-based segmentation~\citep{Truong_2020}, the total cost of splitting at $c$ is
\begin{equation}
 J(c)=E(0,c)+E(c,M),
 \label{eq:sweet-spot-choice}
\end{equation}
and the sweet-spot index is $n_i=\arg\min_{1\le c<M}J(c)$, with ties resolved toward the earliest index. The robot observes after completing $a_{i,n_i}$, the $(n_i+1)$th action. Observing at the end of the first fitted motion segment provides fresh feedback before the chunk finishes without replanning in the middle of a coherent motion, and it leaves the remaining actions to overlap with shared inference. Prefix sums evaluate all candidates in $O(H_i)$ time. Appendix~A details the features and rationale for this boundary, and Table~\ref{tab:sweet-point-success} shows that it achieves the highest observed success in 11 of 12 model--task settings.

\paragraph{Deriving the execution slack.}
The unexecuted suffix translates the trigger into execution slack. With no outstanding request, the robot registers at the first action boundary after the sweet-spot action has been completed or skipped. At registration time $r_i$, it captures the observation and records $p_i$, the index of the next action. With $H_i-p_i$ actions remaining at a constant duration $\Delta_i$ per action, the robot's execution slack is $(H_i-p_i)\Delta_i$ and its modeled buffer-depletion deadline is
\begin{equation}
 D_i=r_i+(H_i-p_i)\Delta_i.
 \label{eq:robot-buffer-deadline}
\end{equation}
The request carries the observation, adapter and generation IDs, and this slack, which the server converts to a deadline on its own clock after accounting for upload time.

\paragraph{Continuing execution across chunks.}
The robot keeps executing buffered actions during communication, queueing, and inference. When the returned chunk $A_i^{\mathrm{new}}$ arrives, it finishes any action in progress and atomically replaces its buffer at handoff time $\tau_i$. Let $s$ count the actions completed since observation, excluding buffer-empty waiting. The robot discards the first $s$ actions of the new chunk and installs the suffix
\begin{equation}
 A_i^{\rm buffer}\leftarrow A_i^{\mathrm{new}}[s:H_i],
 \label{eq:action-handoff}
\end{equation}
which it executes with the new chunk's per-action duration $\Delta_i^{\mathrm{new}}$. The returned sweet-spot index $n_i^{\mathrm{new}}$ then determines when to request again, as shown in Figure~\ref{fig:sweet-spot-execution}. If $s\le n_i^{\mathrm{new}}$, the robot executes through the sweet-spot action before requesting (panel a). If $n_i^{\mathrm{new}}<s<H_i$, that action has already been skipped, so the robot requests immediately after buffer replacement and acknowledgment (panel b). In both cases, the retained suffix depletes at $D_i^{\mathrm{new}}=\tau_i+(H_i-s)\Delta_i^{\mathrm{new}}$, where $H_i-s$ is the retained action count.
\begin{figure}[t]
 \centering
 \includegraphics[width=\columnwidth]{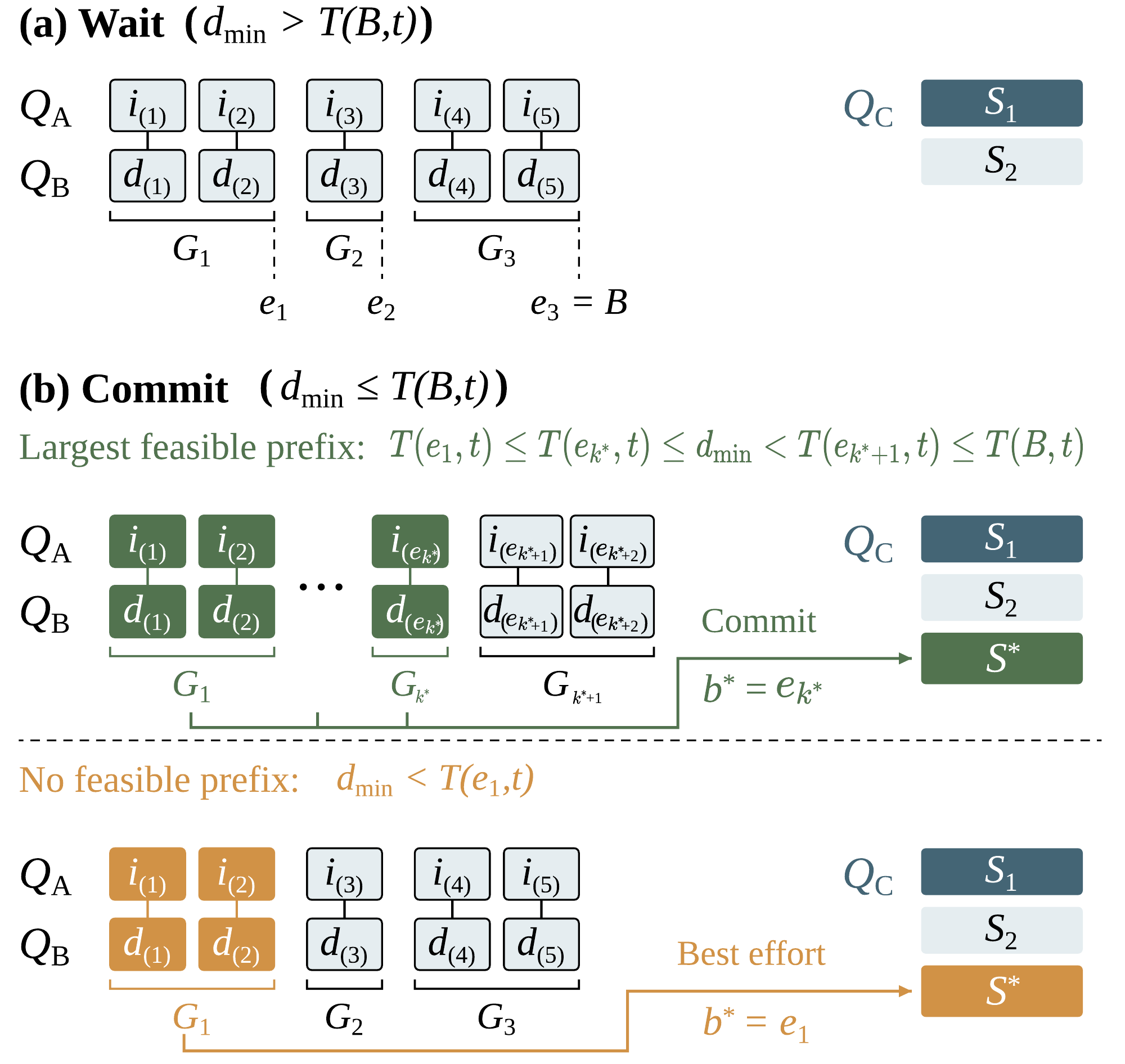}
 \caption{SlackBatch decisions across the three queues. (a) Requests remain in $Q_A/Q_B$ while the full candidate retains a positive dispatch margin. (b) Once the margin is exhausted, commitment moves the largest feasible complete-group prefix to $Q_C$ (green), or the head group if no prefix fits (orange). Unselected requests remain pending.}
 \label{fig:batch-cases}
\end{figure}
\subsection{Slack-Driven Asynchronous Batching}
\label{sec:slack-aware-batching}

As outlined in Figure~\ref{fig:system-overview}, SlackBatch uses the deadlines from MotionExec to decide when to commit a batch and which requests to include. Pending requests and their budgets are paired in $Q_A$ and $Q_B$, and the FIFO queue $Q_C$ holds committed batches that have not started running. At server time $t$, robot $i$ has GPU-completion budget $d_i(t)=D_i-t-\delta_i$, where $D_i$ is its buffer-depletion deadline on the server clock and $\delta_i$ is the offline-profiled latency from GPU completion to result receipt. Pending requests are ordered by increasing budget, with ties broken by registration order. Let $d_{(j)}(t)$ denote the budget at position $j$, so the head has the minimum budget $d_{\min}(t)=d_{(1)}(t)$.

\paragraph{Estimating GPU time.}
For a candidate batch of $b$ requests, the predicted GPU time includes preceding work and its own inference:
\begin{equation}
 T(b,t)=\underbrace{\widehat R(t)+\sum_{S\in Q_C}\widehat\ell(|S|)}_{W(t)}+\widehat\ell(b).
 \label{eq:batch-response-time}
\end{equation}
Here $\widehat\ell(b)$ is the offline-profiled, nondecreasing inference time for $b$ requests, and $W(t)$ is the work of the running batch and the queued batches $S\in Q_C$. For a running batch of size $b_a$ started at $t_a$, $\widehat R(t)=\max\{0,\widehat\ell(b_a)-(t-t_a)\}$, or zero when the GPU is idle. The dispatch margin of a $b$-request candidate is $d_{\min}(t)-T(b,t)$.

\paragraph{Forming and selecting batches.}
SlackBatch groups adjacent requests whenever $d_{(j+1)}(t)-d_{(j)}(t)<\widehat\ell(1)$, yielding $m$ indivisible groups $G_1,\ldots,G_m$. Serving the earlier request of such a pair alone would occupy the GPU for $\widehat\ell(1)$ and leave the later one with less budget than the earlier one, so grouping keeps similarly urgent requests in one round and favors larger batches. Valid batch sizes are $e_k=\sum_{h=1}^{k}|G_h|$ for $1\le k\le m$, and the largest candidate $B=e_m$ includes all pending requests, capped at $B_{\max}=N$.

As Figure~\ref{fig:batch-cases}(a) shows, SlackBatch waits while the full candidate retains a positive dispatch margin ($d_{\min}(t)>T(B,t)$) and sets a waiting expiry at $t+d_{\min}(t)-T(B,t)$. Once the margin is exhausted, it commits the largest feasible complete-group prefix, or the head group if none fits:
\begin{equation}
 b^\star=\max\bigl(\{e_1\}\cup\{e_k:T(e_k,t)\le d_{\min}(t)\}\bigr).
 \label{eq:feasible-batch-sizes}
\end{equation}
Feasibility depends only on the head because every other member has at least as much budget, and $e_1$ provides the best-effort case in Figure~\ref{fig:batch-cases}(b) when even the head group misses $d_{\min}(t)$. The selected $k^\star$ groups form batch $S^\star$ with $b^\star=e_{k^\star}$ requests. Commitment atomically moves $S^\star$ from $Q_A/Q_B$ to $Q_C$ and adds $\widehat\ell(b^\star)$ to $W(t)$ before the remaining requests are reevaluated.

\paragraph{Updating estimates.}
Between events, budgets and $\widehat R(t)$ decrease with elapsed time. At each GPU completion, the runtime starts the next batch from $Q_C$, and SlackBatch recomputes $\widehat R$ and $W$ from actual progress while $\widehat\ell$ remains fixed. SlackBatch is invoked on request arrival, GPU completion, and waiting expiry, the last detected by a nominal 5\,ms poll. Because all budgets decrease at the same rate and the head group is committed whenever the margin is exhausted, every pending request eventually reaches the head, so SlackBatch defers requests with large slack without starving them. Appendix~B gives the complete algorithm.

\begin{figure}[t]
 \centering
 \includegraphics[width=\columnwidth]{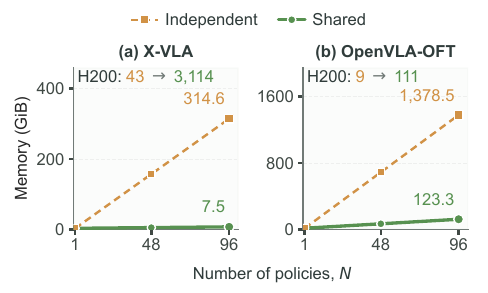}
 \caption{Memory savings from backbone sharing. Model-weight memory for 1--96 policies under independent deployment with merged checkpoints and shared multi-LoRA deployment: (a) X-VLA and (b) OpenVLA-OFT. H200 annotations indicate the model-weight-based resident policy capacity under independent and shared deployment, respectively.}
 \label{fig:resource-scaling}
\end{figure}
\section{Evaluation}
\label{sec:new-evaluation}

\subsection{Experimental Setup}
\label{sec:new-eval-setup}

\paragraph{Experimental setting.}
We evaluate MotionServe on a server with four NVIDIA H200 SXM GPUs (141\,GB each), using one GPU for shared inference and three for robot simulation. We serve X-VLA~\citep{ICLR2026_630d2938} and OpenVLA-OFT~\citep{KimM1-RSS-25}, whose action-chunk lengths are 30 and 8. Each task group has its own LoRA adapter, with rank 8 for X-VLA and 64 for OpenVLA-OFT. Each robot is served through its own adapter instance, so a fleet of $N$ robots presents $N$ resident adapters, and serving cost depends on the adapter count rather than adapter values. The runtime is built on PyTorch 2.11 and executes mixed-adapter batches with fused SGMV LoRA kernels~\citep{MLSYS2024_054de805} in BF16. With one inference GPU holding at most 111 OpenVLA-OFT policies (Section~\ref{sec:new-eval-memory}), we test $N\in\{8,16,32,64,96\}$ to keep all policies resident.

\paragraph{Tasks and workloads.}
X-VLA runs four LIBERO suites with a simulated Franka Panda arm~\citep{NEURIPS2023_8c3c6668} and four SimplerEnv tasks with a simulated WidowX-250 arm~\citep{pmlr-v270-li25c}, and OpenVLA-OFT runs the four LIBERO suites (see Appendix~C for robot--task assignments and per-robot policy details). Each robot samples a nominal action period uniformly from 0.1--2.0\,s per episode, spanning reported robot-arm control rates~\citep{chi2024diffusionpolicy,10611548}, and perturbs it for each new chunk following action-timestep randomization~\citep{Sim2Real2018}. Simulation runs in real time, so physical execution, network transfer, and GPU serving share one wall clock. 
\begin{figure*}[t]
 \centering
 \includegraphics[width=\textwidth]{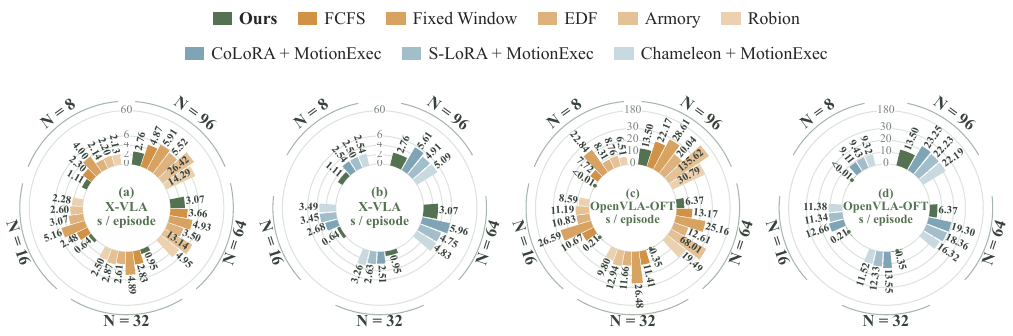}
 \caption{\textbf{Mean robot stall time (s/episode).} X-VLA (a,b) and OpenVLA-OFT (c,d), averaged over four seeds. Orange compares robot-serving baselines, blue compares adapted multi-LoRA batching with MotionExec, and green denotes MotionServe. Sectors indicate fleet size. }
 \label{fig:robot-stall-comparison}
\end{figure*}

\paragraph{Baselines.}
We compare MotionServe with eight baselines in three categories. \emph{Conventional multi-robot scheduling.} first-come-first-served (FCFS), Fixed Window~\citep{201468}, and earliest-deadline-first (EDF) use request arrival order, bounded batching delays, and deadline order, respectively. \emph{Robot-specific serving designs.} Armory~\citep{bansal2026actionchunkschedulingbatched} uses Lookahead scheduling based on modeled action-buffer evolution, and Robion~\citep{adamopoulos2026efficientvisionlanguageactionmanagementserving} combines SLO-aware request selection with concurrent VLA-stage execution. \emph{Multi-LoRA scheduling ports.} CoLoRA~\citep{lin2026colora}, S-LoRA~\citep{MLSYS2024_906419cd}, and Chameleon~\citep{Iliakopoulou_2025} supply scheduling and admission mechanisms adapted to complete VLA requests. Unlike the baselines above, they share MotionExec's observation and handoff mechanism, so scheduling is the only difference from MotionServe. All configurations use the same batch-size cap $B_{\max}=N$, and Appendix~D.2 details the runtime adaptation of each baseline.

\paragraph{Metrics.}
We report memory footprint, resident policy count, mean robot stall, task success, and throughput. Stall accumulates from buffer exhaustion until new actions become executable, and mean stall averages it over all evaluated episodes, including failed and step-limit-truncated ones. Task success is measured over 500 episodes per LIBERO suite or Simpler task in each seed. Each condition runs with four seeds, and we report the mean of the seed-level values. 

Appendix~D provides details of the experimental setup.

\subsection{Memory Savings and Policy Capacity}
\label{sec:new-eval-memory}
Following Equation~\eqref{eq:bg-memory}, independent deployment stores a complete VLA checkpoint for each policy, whereas shared deployment stores the backbone once and adds only policy-specific parameters for each policy. Figure~\ref{fig:resource-scaling} compares the two deployments. At 96 policies, shared deployment reduces weight memory by 97.6\% for X-VLA and 91.1\% for OpenVLA-OFT, to 7.52 and 123.35\,GiB, respectively. On one H200 SXM, independent deployment fits only 43 X-VLA or 9 OpenVLA-OFT policies, whereas shared deployment fits 3{,}114 and 111. This capacity lets one H200 keep every policy of the 96-robot fleets in the following experiments resident, avoiding adapter loading during serving.

\begin{figure}[t]
 \centering
 \includegraphics[width=\columnwidth]{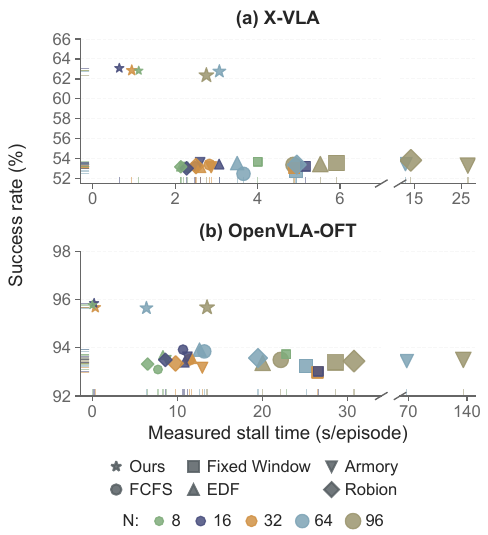}
 \caption{Task success versus robot stall. (a) X-VLA on LIBERO and WidowX. (b) OpenVLA-OFT on LIBERO.  Shapes denote methods (Ours: MotionServe). Color and size denote $N$. Upper left is better. Broken horizontal axes use different linear scales.}
 \label{fig:robot-success-stall}
\end{figure}

\subsection{Robot Stall Time}
\label{sec:new-eval-stall}

Figure~\ref{fig:robot-stall-comparison} compares MotionServe (green) with conventional and robot-specific serving baselines (orange), and multi-LoRA batching policies paired with MotionExec (blue). Among these configurations, MotionServe achieves the lowest mean stall for both models at every tested load, and its advantage persists as the fleet grows from 8 to 96 robots. The orange panels (a,c) compare MotionServe with conventional multi-robot schedulers (FCFS, Fixed Window, and EDF) and robot-specific serving systems (Armory and Robion). At $N=96$, MotionServe reduces X-VLA's mean stall from 4.87\,s per episode with FCFS, the lowest-stall baseline in these categories, to 2.76\,s, a 43.4\% reduction. For OpenVLA-OFT, it reduces mean stall from 20.04\,s with EDF, the lowest-stall baseline, to 13.50\,s, a 32.6\% reduction. Section~\ref{sec:ablation-study} separates the contributions of MotionExec and SlackBatch to these gains.

The blue panels (b,d) hold MotionExec fixed and compare SlackBatch with the batching policies adapted from CoLoRA, S-LoRA, and Chameleon. At $N=96$, these policies incur 4.91--5.61\,s of stall per episode on X-VLA and 22.19--23.25\,s on OpenVLA-OFT. SlackBatch reduces stall by 43.8\% relative to S-LoRA on X-VLA and by 39.2\% relative to Chameleon on OpenVLA-OFT, the lowest-stall adapted policy for each model. These results show that existing multi-LoRA batching policies do not transfer directly to multi-robot serving. SlackBatch differs by using the dispatch margin to decide how long to defer dispatch and which requests to batch, which lowers stall for both models. Beyond these mean improvements, Appendix~E shows that MotionServe also achieves the lowest median stall at every fleet size for both models, indicating that its gains extend across the fleet rather than coming from a few robots.

\subsection{Task Success--Stall Trade-off}

\label{sec:new-eval-success-stall}

Figure~\ref{fig:robot-success-stall} plots task success against mean robot stall for MotionServe, the conventional multi-robot schedulers (FCFS, Fixed Window, and EDF), and the robot-specific serving systems (Armory and Robion). MotionServe's points lie in the upper-left region of both panels at every tested fleet size, so it improves both metrics rather than trading one for the other. At $N=96$, MotionServe achieves 62.35\% mean success on X-VLA across LIBERO and WidowX and 95.68\% on OpenVLA-OFT, with mean stalls of 2.76 and 13.50\,s per episode, respectively. Compared with FCFS on X-VLA and EDF on OpenVLA-OFT, the lowest-stall baseline for each model, it improves success by 8.98 and 2.30 percentage points and reduces stall by 43.4\% and 32.6\%. Section~\ref{sec:ablation-study} shows that MotionExec drives most of this success gain on X-VLA, while SlackBatch adds further stall reduction.
\begin{table}[t]
 \centering\small
 \caption{Task throughput (tasks/hour). X-VLA covers LIBERO and WidowX. OpenVLA-OFT covers LIBERO. Bold marks the highest throughput at each $N$.}
 \label{tab:task-throughput}
\vskip 0.15in
 \setlength{\tabcolsep}{3pt}
 \renewcommand{\arraystretch}{1.0}
 \setlength{\aboverulesep}{1.5pt}
 \setlength{\belowrulesep}{1.5pt}
 \begin{tabular*}{\columnwidth}{@{\extracolsep{\fill}}lccccc@{}}
  \toprule
  \textbf{X-VLA} & $N=8$ & $16$ & $32$ & $64$ & $96$ \\
  \midrule
  \textbf{Ours} & \textbf{196} & \textbf{408} & \textbf{765} & \textbf{1,408} & \textbf{2,197} \\
  FCFS & 163 & 338 & 637 & 1,171 & 1,841 \\
  Fixed Window & 162 & 328 & 620 & 1,162 & 1,828 \\
  EDF & 164 & 337 & 635 & 1,196 & 1,831 \\
  Armory & 164 & 340 & 633 & 1,092 & 1,513 \\
  Robion & 164 & 337 & 637 & 1,176 & 1,697 \\
  \midrule
  \textbf{OpenVLA-OFT} & $N=8$ & $16$ & $32$ & $64$ & $96$ \\
  \midrule
  \textbf{Ours} & \textbf{82} & \textbf{151} & \textbf{294} & \textbf{601} & \textbf{900} \\
  FCFS & 78 & 144 & 279 & 579 & 859 \\
  Fixed Window & 75 & 137 & 267 & 557 & 844 \\
  EDF & 78 & 143 & 279 & 580 & 863 \\
  Armory & 78 & 143 & 277 & 502 & 660 \\
  Robion & 78 & 144 & 280 & 567 & 840 \\
  \bottomrule
 \end{tabular*}
\end{table}

\begin{table}[t]
\centering\small
\setlength{\tabcolsep}{3pt}
\renewcommand{\arraystretch}{1.0}
\setlength{\aboverulesep}{1.5pt}
\setlength{\belowrulesep}{1.5pt}
\caption{Task success (\%) by observation trigger. All triggers share the same overlap-and-skip executor. Fixed-prefix triggers observe after $p$ actions, and $n$ observes at the sweet spot. The largest prefixes, $p=30$ and $p=8$, observe after the full chunk. Bold marks the best result for each model and task.}
\label{tab:sweet-point-success}
\vskip 0.15in
\begin{tabular*}{\columnwidth}{@{\extracolsep{\fill}}lrrrr@{}}
\toprule
\textbf{X-VLA} & $p=5$ & $p=10$ & $p=30$ & $n$ \\
\midrule
\multicolumn{5}{@{}l}{\textit{LIBERO}} \\
Spatial & 93.20 & \textbf{93.60} & 90.80 & 93.40 \\
Object & 41.20 & 90.80 & 96.80 & \textbf{97.60} \\
Goal & 71.60 & 90.60 & 95.00 & \textbf{95.80} \\
Long & 51.40 & 80.60 & 80.40 & \textbf{85.40} \\
\addlinespace[2pt]
\multicolumn{5}{@{}l}{\textit{Simpler / WidowX}} \\
Stack Cube & 0.00 & 8.50 & 13.00 & \textbf{17.50} \\
Carrot & 0.00 & 13.50 & 18.00 & \textbf{25.50} \\
Eggplant & 0.00 & 34.00 & 25.50 & \textbf{59.00} \\
Spoon & 4.50 & 13.50 & 12.50 & \textbf{29.00} \\
\midrule
\textbf{OpenVLA-OFT} & $p=2$ & $p=3$ & $p=8$ & $n$ \\
\midrule
\multicolumn{5}{@{}l}{\textit{LIBERO}} \\
Spatial & 90.80 & 91.40 & 91.60 & \textbf{92.00} \\
Object & 88.20 & 93.40 & 98.20 & \textbf{98.60} \\
Goal & 91.20 & 92.00 & 96.60 & \textbf{98.00} \\
Long & 69.80 & 83.00 & 93.40 & \textbf{94.00} \\
\bottomrule
\end{tabular*}
\end{table}

\subsection{Task Throughput}
\label{sec:new-eval-throughput}

Task throughput counts successful task completions across all robots per hour of wall-clock runtime, measured over the same run interval. Table~\ref{tab:task-throughput} compares MotionServe with the end-to-end baselines of Section~\ref{sec:new-eval-stall} at $N\in\{8,16,32,64,96\}$. MotionServe achieves the highest throughput for both models at every fleet size, improving on the highest-throughput baseline at each $N$ by 17.7--20.2\% on X-VLA and 3.6--5.1\% on OpenVLA-OFT. At $N=96$, it completes 2,197 tasks per hour on X-VLA and 900 on OpenVLA-OFT, 19.3\% and 4.3\% more than FCFS and EDF, respectively. Its gains over Armory and Robion at the same load reach 45.2\% and 29.5\% on X-VLA and 36.3\% and 7.2\% on OpenVLA-OFT. In deployment terms, the same GPU completes 4--20\% more successful tasks per hour without additional hardware.

\subsection{Effect of Sweet-Spot Observation Timing}
\label{sec:new-eval-sweet-point}

Table~\ref{tab:sweet-point-success} compares fixed-prefix triggers with sweet-spot observation under the same overlap-and-skip protocol, differing only in observation timing. Observing too early can reduce success, as with X-VLA at $p=5$ on LIBERO-Object (41.20\%), whereas extending the prefix does not consistently help, as with X-VLA on WidowX Eggplant (34.00\% at $p=10$ versus 25.50\% at $p=30$). Sweet-spot observation achieves the highest success in 11 of 12 model--task settings and outperforms full-chunk observation ($p=30$ for X-VLA and $p=8$ for OpenVLA-OFT) on every task. Even against the post-hoc best fixed-prefix trigger for each task, it raises mean success for both models, by up to 7.34 percentage points. The largest per-task gain reaches 25.0 percentage points on WidowX Eggplant with X-VLA. These results show that motion-guided observation observes as early as the predicted motion allows without sacrificing success, leaving buffered actions as execution slack for shared inference and batching.
\begin{table}[t]
 \centering\small
 \caption{Component ablations on X-VLA. Results cover LIBERO and WidowX with $B_{\max}=N$. Paired rows show each baseline without (--) and with the component listed under \emph{Added}. Bold marks the best result in each column. Appendix~D.2 details the paired controls.}
 \label{tab:ablation-study}
 \vskip 0.15in
 \setlength{\tabcolsep}{1.25pt}
 \renewcommand{\arraystretch}{1.0}
 \setlength{\aboverulesep}{2.5pt}
 \setlength{\belowrulesep}{2.5pt}
 \begin{tabular*}{\columnwidth}{@{\extracolsep{\fill}}l@{\hspace{4pt}}lrrrrr@{}}
  \toprule
  \multicolumn{7}{@{}l}{\textbf{Success rate (\%)} $\uparrow$} \\
  Method & Added & $N=8$ & $16$ & $32$ & $64$ & $96$ \\
  \midrule
  FCFS & -- & 53.03 & 53.23 & 53.40 & 52.48 & 53.38 \\
  FCFS & MotionExec & 61.62 & 61.49 & 61.37 & 61.13 & 62.21 \\
  \addlinespace[2pt]
  Fixed Window & -- & 53.70 & 53.25 & 53.05 & 52.75 & 53.55 \\
  Fixed Window & MotionExec & 60.85 & 62.18 & 61.52 & 60.71 & 60.93 \\
  \addlinespace[2pt]
  EDF & -- & 53.20 & 53.45 & 53.15 & 53.55 & 53.45 \\
  EDF & MotionExec & 61.25 & 61.23 & 60.78 & 61.42 & 60.66 \\
  \midrule
  PACE & -- & 57.03 & 61.50 & 60.73 & 57.90 & 58.35 \\
  PACE & SlackBatch & 57.60 & 60.04 & 60.97 & 59.04 & 58.13 \\
  \addlinespace[2pt]
  AAC & -- & 55.08 & 54.53 & 55.13 & 53.07 & 55.67 \\
  AAC & SlackBatch & 54.40 & 56.48 & 53.98 & 54.06 & 54.83 \\
  \addlinespace[2pt]
  AutoHorizon & -- & 58.45 & 58.58 & 59.26 & 58.15 & 58.95 \\
  AutoHorizon & SlackBatch & 57.65 & 59.63 & 58.16 & 58.82 & 58.27 \\
  \midrule
  \multicolumn{2}{@{}l}{\textbf{MotionServe (Ours)}} & \textbf{62.83} & \textbf{63.08} & \textbf{62.85} & \textbf{62.75} & \textbf{62.35} \\
  \midrule[0.6pt]
  \multicolumn{7}{@{}l}{\textbf{Mean stall (s/episode)} $\downarrow$} \\
  Method & Added & $N=8$ & $16$ & $32$ & $64$ & $96$ \\
  \midrule
  FCFS & -- & 2.30 & 2.48 & 2.83 & 3.66 & 4.87 \\
  FCFS & MotionExec & \textbf{1.07} & 0.69 & 1.08 & 3.39 & 3.47 \\
  \addlinespace[2pt]
  Fixed Window & -- & 4.00 & 5.16 & 4.89 & 4.93 & 5.91 \\
  Fixed Window & MotionExec & 1.13 & 0.65 & 1.52 & 3.84 & 5.12 \\
  \addlinespace[2pt]
  EDF & -- & 2.14 & 3.07 & 2.61 & 3.50 & 5.52 \\
  EDF & MotionExec & 1.14 & 0.82 & 2.01 & 3.21 & 3.30 \\
  \midrule
  PACE & -- & 6.07 & 4.71 & 7.22 & 10.13 & 15.53 \\
  PACE & SlackBatch & 2.31 & 2.39 & 2.24 & 3.77 & 5.53 \\
  \addlinespace[2pt]
  AAC & -- & 9.61 & 10.53 & 20.45 & 63.06 & 118.91 \\
  AAC & SlackBatch & 9.22 & 10.39 & 18.65 & 57.33 & 105.33 \\
  \addlinespace[2pt]
  AutoHorizon & -- & 4.63 & 5.07 & 5.62 & 8.03 & 11.81 \\
  AutoHorizon & SlackBatch & 2.84 & 3.01 & 2.98 & 4.48 & 9.54 \\
  \midrule
  \multicolumn{2}{@{}l}{\textbf{MotionServe (Ours)}} & 1.11 & \textbf{0.64} & \textbf{0.95} & \textbf{3.07} & \textbf{2.76} \\
  \bottomrule
 \end{tabular*}
\end{table}

\subsection{Ablation Study}
\label{sec:ablation-study}

Table~\ref{tab:ablation-study} reports paired X-VLA configurations across LIBERO and WidowX tasks, with $B_{\max}=N$ throughout. For FCFS, Fixed Window, and EDF, each pair fixes the scheduler and pipelined wrapper while replacing the fixed observation trigger with MotionExec's motion-based trigger (Appendix~D.2). MotionExec improves success by 7.15--8.93 percentage points and reduces mean stall at every tested fleet size, by 53.5\%, 71.8\%, and 46.7\% at $N=8$ for FCFS, Fixed Window, and EDF, respectively.

To test SlackBatch with other observation triggers, we integrate the execution-horizon selectors PACE~\citep{nie2026pace}, AAC~\citep{Liang_2026_CVPR}, and AutoHorizon~\citep{10.1007/978-3-032-37196-6_33} into the common pipelined action executor. Each FCFS-versus-SlackBatch pair retains the same selector, wrapper, and inference work, including AAC's multiple samples and AutoHorizon's attention extraction, so only scheduling changes (Appendix~D.2). Across the tested fleet sizes, SlackBatch reduces mean stall by 49.3--69.0\% for PACE, 19.2--47.0\% for AutoHorizon, and 1.3--11.4\% for AAC, while keeping success within 2 percentage points of its FCFS counterpart. AAC gains less, because its long selected prefixes leave few buffered actions to overlap service and its 20-sample entropy estimation adds inference work. MotionServe combines both components and achieves the highest success rate at every tested fleet size and the lowest mean stall at $N\ge16$. At $N=8$, its mean stall of 1.11\,s per episode is close to FCFS+MotionExec's 1.07\,s, while its success rate is higher (62.83\% versus 61.62\%). At $N=96$, it reduces mean stall to 2.76\,s per episode, 20.5\% and 16.4\% below FCFS+MotionExec and EDF+MotionExec, respectively. MotionExec thus drives the success gain and part of the stall reduction, while SlackBatch adds stall reduction that matters most at scale.

\section{Conclusion}
\label{sec:conclusion}

MotionServe serves many task-specific VLA policies on a shared GPU by treating each robot's execution slack as the link between when robots observe and how the server batches their requests. MotionExec times each observation from the predicted motion, leaving buffered actions as execution slack, and SlackBatch spends this slack on batch formation by deferring dispatch only while the dispatch margin remains positive. MotionServe allows a single GPU to host all policies, and coordinating observation timing with batching reduces mean stall by 32.6--43.4\% while improving task success and throughput. These results demonstrate the value of connecting pipelined action execution with multi-LoRA serving and establish a practical approach to deploying diverse robot policies on shared accelerators.

\bibliography{reference}
\bibliographystyle{mlsys2025}

\section*{AI Usage Disclosure}

In this work, we used ChatGPT (OpenAI) to assist
with language editing, grammar checking, and
software code editing. All AI-assisted suggestions
were reviewed by the authors before incorporation.
We take full responsibility for the final content
of this work, including text, code, experimental
results, and claims.

\input{appendix}
\end{document}

%% file: appendix.tex
\onecolumn

\fancyhead[C]{\small\bfseries MotionServe: Efficient Multi-Robot Serving --- Appendix}
\thispagestyle{empty}
\begin{center}
{\large\bfseries MotionServe: Efficient Multi-Robot Serving by Exploiting Robot Execution Slack\par}
\smallskip
{\normalsize\bfseries Appendix\par}
\end{center}

\appendix

\section{Rationale for Sweet-Spot Selection}
\label{sec:sweet-spot-rationale}

\paragraph{Observation timing and serving slack.}
The observation trigger determines both the state supplied to the next prediction and the time available to process it. After completing $a_c$ in a chunk executed from its first action, the robot retains $H_i-c-1$ actions, yielding an execution budget of $(H_i-c-1)\Delta_i$. Earlier capture increases this budget at a fixed action period. A fixed prefix, however, places the trigger at the same index regardless of the predicted motion. MotionExec instead uses motion structure as a proxy for execution progress. Its premise is that a transition in motion trend or gripper command provides a useful point to update feedback for subsequent actions. The generated chunk supplies this signal without an additional policy inference.

\paragraph{Motion and interaction features.}
Omitting the robot index, a policy-specific decoder represents action $a_j$ by end-effector position $q_j$, orientation $R_j$, and binary gripper command $g_j$. For the $M=H_i-1$ motion intervals, the normalized features are
\begin{equation}
 z_j=\left[\frac{q_{j+1}-q_j}{\sigma_p},\thickspace
 \frac{\operatorname{Log}(R_{j+1}R_j^\top)}{\sigma_R}\right],
 \label{eq:sweet-spot-features}
\end{equation}
where $0\le j<M$ and $\operatorname{Log}$ returns the principal rotation vector. The fixed scales $\sigma_p=0.01\,\mathrm{m}$ and $\sigma_R=0.10\,\mathrm{rad}$ make the residuals dimensionless and determine the relative weighting of the two motion components. The affine model $\alpha+\beta j$ captures both constant increments and gradual linear variation within a segment, so the segmentation criterion responds to changes that one such trend explains poorly.

Gripper commands complement this geometric description with a discrete interaction signal. In Equation~2, $G(0,r)=\{0,\ldots,r\}$ and $G(l,r)=\{l+1,\ldots,r\}$ for $l>0$, so every command is assigned to exactly one of the two segments. For binary commands, the constant-fit residual is $|G(l,r)|\bar g(1-\bar g)$, where $\bar g$ is their mean over $G(l,r)$. The residual vanishes for a consistent command and penalizes mixtures of open and closed commands. The coefficient 4 sets its weight relative to the normalized motion residual. These terms jointly favor segments with internally consistent motion trends and gripper commands.

\paragraph{Single-boundary objective.}
Each request cycle requires one next observation trigger. We therefore use a fixed two-segment objective~\citep{Truong_2020}, fitting independent motion trends and gripper means on either side of each candidate $c$. All candidates account for the same motion intervals and gripper commands, each exactly once. A misplaced boundary can leave different trends within one segment and increase its fitting error. Minimizing $J(c)=E(0,c)+E(c,M)$ selects the boundary that best supports separate descriptions of the two portions. Equivalently, it maximizes the improvement $E(0,M)-J(c)$ over fitting the entire chunk as one segment. This improvement is nonnegative because both segment fits can reuse the unsplit parameters. The objective thus measures how much a boundary improves the description of predicted motion. We use this structural criterion as a surrogate for observation timing, with its effect on task completion evaluated in Table~3.

\paragraph{Earliest observation relative to the selected boundary.}
MotionExec uses completion of the first fitted segment as a proxy for sufficient execution progress. The selected index $n_i=c$ places the end of that segment at action $a_c$. For a chunk executed from its first action, an observation captured before $a_c$ completes precedes completion of this fitted prefix. Observing immediately after $a_c$ completes is therefore the earliest action-aligned observation that satisfies this progress criterion. Waiting until a later action $a_k$, with $c<k<H_i$, consumes an additional $(k-c)\Delta_i$ of execution slack without being required to complete that prefix. This gives the timing rationale for observing at the inferred boundary while retaining the remaining actions for inference and batching.

Boundary selection follows Equation~3, with earliest-index tie-breaking preserving the most execution slack among candidates with minimum fitting cost. Thus, earliest refers to the selected motion-derived progress criterion. The runtime triggers at the first eligible action boundary after the selected action has completed or been skipped, as detailed in Section~3.2, and subsequent handoff follows Equation~5. Because the two-segment fit also returns a boundary when no distinct transition exists, the criterion remains a deterministic motion-based heuristic. Prefix sums evaluate it in $O(H_i)$ time, and Table~3 assesses its task-level effectiveness.
\newpage
\section{SlackBatch Scheduling Algorithm}
\label{sec:slackbatch-algorithm}
Algorithm~\ref{alg:iterative-batching} runs on request arrival, GPU completion, or timer expiry. It refreshes budgets and GPU work, then either waits or commits a complete-group prefix according to Equations~6 and~7. Deadlines $D_i$ follow Equation~4, and $\delta_i$ is the offline result-delivery latency. The sets $\mathcal E$ and $\mathcal F$ denote legal prefix sizes and their feasible subset. Queue slices use one-based inclusive indices, and $\Vert$ denotes concatenation. The tuple update atomically removes selected request--budget pairs and appends their batch to $Q_C$. \textsc{WakeExecutor} starts the FIFO head when the GPU is idle. With at most one outstanding request per robot, $B\le B_{\max}=N$, where $N$ is the number of robots.
\begin{algorithm}[!ht]
 \caption{SlackBatch}
 \label{alg:iterative-batching}
 \renewcommand{\algorithmicrequire}{\textbf{Input:}}
 \renewcommand{\algorithmiccomment}[1]{\hfill{\footnotesize$\triangleright$ #1}}
 \begin{algorithmic}[1]
 \REQUIRE Pending requests $Q_A$, paired budgets $Q_B$, committed FIFO $Q_C$, executor state
 \REQUIRE Buffer deadlines $\{D_i\}$, delivery estimates $\{\delta_i\}$, inference profile $\widehat\ell$
 \STATE $\textsc{CancelTimer}()$
 \WHILE{$Q_A\ne\varnothing$}
  \STATE $t\gets\textsc{Now}(),\quad B\gets|Q_A|$
  \STATE $Q_B\gets[D_i-t-\delta_i]_{i\in Q_A}$
  \STATE $(Q_A,Q_B)\gets\textsc{Sort}(Q_A,Q_B)$ \COMMENT{Budget, then registration order}
  \STATE $(d_{(1)},\ldots,d_{(B)})\gets Q_B$
  \STATE $W\gets\widehat R(t)+\sum_{S\in Q_C}\widehat\ell(|S|)$ \COMMENT{Equation~6}
  \STATE $\mathcal E\gets\{j:1\le j<B,\ d_{(j+1)}-d_{(j)}\ge\widehat\ell(1)\}\cup\{B\}$ \COMMENT{Group boundaries}
  \IF{$d_{(1)}>W+\widehat\ell(B)$}
   \STATE $\textsc{ArmTimer}\bigl(t+d_{(1)}-W-\widehat\ell(B)\bigr)$
   \STATE \textbf{return}
  \ENDIF
  \STATE $\mathcal F\gets\{b\in\mathcal E:W+\widehat\ell(b)\le d_{(1)}\}$
  \IF{$\mathcal F=\varnothing$}
   \STATE $b^\star\gets\min\mathcal E$ \COMMENT{Best effort}
  \ELSE
   \STATE $b^\star\gets\max\mathcal F$
  \ENDIF
  \STATE $S^\star\gets Q_A[1:b^\star]$
  \STATE $(Q_A,Q_B,Q_C)\gets\bigl(Q_A[b^\star+1:B],\ Q_B[b^\star+1:B],\ Q_C\mathbin{\Vert}[S^\star]\bigr)$ \COMMENT{Atomic}
  \STATE $\textsc{WakeExecutor}()$
 \ENDWHILE
 \end{algorithmic}
\end{algorithm}

\newpage

\section{Robot--Task Assignment and Per-Robot Policies}
\label{sec:task-coverage}

MotionServe and all eight baselines use the same robot--task mapping for each model and robot count. X-VLA evenly distributes robots across four LIBERO suites and four SimplerEnv WidowX tasks, assigning $N/8$ robots to each. OpenVLA-OFT uses only the four LIBERO suites, with its two task groups per suite together containing $N/4$ robots. Table~\ref{tab:task-coverage} reports the number of robots assigned to each suite or task at each total robot count $N$.

Each robot uses a dedicated LoRA adapter indexed by its robot ID, even when several robots execute the same benchmark task. Requests from those robots therefore select separate adapters on the shared server. This setup emulates a fleet of $N$ robot-specific task policies with $N$ independent LoRA adapters, while all serving methods retain the same task assignment.

\begin{table}[!ht]
\centering\small
\caption{Robot allocation by task suite or task for X-VLA and OpenVLA-OFT. Entries give the number of assigned robots, and $N$ is the total robot count per model. A dash denotes an unused task. All serving methods use the same allocation.}
\label{tab:task-coverage}
\vskip 0.15in
\setlength{\tabcolsep}{3pt}
\renewcommand{\arraystretch}{1.25}
\begin{tabular*}{\textwidth}{@{\extracolsep{\fill}}l*{10}{c}@{}}
\toprule
Task suite or task & \multicolumn{5}{c}{X-VLA} & \multicolumn{5}{c}{OpenVLA-OFT} \\
\cmidrule(lr){2-6}\cmidrule(lr){7-11}
 & $N=8$ & $N=16$ & $N=32$ & $N=64$ & $N=96$ & $N=8$ & $N=16$ & $N=32$ & $N=64$ & $N=96$ \\
\midrule
LIBERO Spatial & 1 & 2 & 4 & 8 & 12 & 2 & 4 & 8 & 16 & 24 \\[2pt]
LIBERO Object & 1 & 2 & 4 & 8 & 12 & 2 & 4 & 8 & 16 & 24 \\[2pt]
LIBERO Goal & 1 & 2 & 4 & 8 & 12 & 2 & 4 & 8 & 16 & 24 \\[2pt]
LIBERO-10 & 1 & 2 & 4 & 8 & 12 & 2 & 4 & 8 & 16 & 24 \\[2pt]
WidowX stack cube & 1 & 2 & 4 & 8 & 12 & --- & --- & --- & --- & --- \\[2pt]
WidowX carrot on plate & 1 & 2 & 4 & 8 & 12 & --- & --- & --- & --- & --- \\[2pt]
WidowX eggplant in basket & 1 & 2 & 4 & 8 & 12 & --- & --- & --- & --- & --- \\[2pt]
WidowX spoon on towel & 1 & 2 & 4 & 8 & 12 & --- & --- & --- & --- & --- \\[2pt]
\bottomrule
\end{tabular*}
\end{table}
\newpage
\section{Implementation Details}
\label{sec:detailed-experimental-setup}

\subsection{Testbed, Policies, and Task Allocation}

The testbed comprises four NVIDIA H200 SXM GPUs, each with 141\,GB of device memory. One GPU hosts the shared VLA inference server and all resident LoRA policies. The other three GPUs run the robot simulations. We evaluate X-VLA and OpenVLA-OFT separately, with action-chunk lengths of 30 and 8. Each robot has a dedicated LoRA policy and sends its adapter ID with each observation. The fleet sizes are $N\in\{8,16,32,64,96\}$. The upper value keeps the per-robot policy count below the 111-policy OpenVLA-OFT residency measured on the single inference GPU, as described in Section~4.2.

X-VLA workloads contain eight task groups: LIBERO Spatial, Object, Goal, and LIBERO-10 with a simulated Franka Panda arm~\citep{NEURIPS2023_8c3c6668}, together with four SimplerEnv tasks using a simulated WidowX-250 6-DoF arm~\citep{pmlr-v270-li25c}. The latter involve stacking a cube, placing a carrot on a plate, putting an eggplant in a basket, and placing a spoon on a towel. OpenVLA-OFT uses eight LIBERO groups, with two groups per suite. For every model--method--load condition, each seed evaluates 500 episodes per LIBERO suite and 500 episodes per Simpler task. OpenVLA-OFT's two groups for a suite share its 500 episodes. The per-seed totals are therefore 4,000 episodes for X-VLA across its four suites and four Simpler tasks, and 2,000 for OpenVLA-OFT across its four suites. Robots execute in closed-loop simulation while observations and action chunks pass through the physical GPU server. Request queueing, inference, result delivery, and action handoff occur during execution.

\paragraph{Repeated runs.}
We repeat each model--method--load condition using four random seeds: 111, 222, 333, and 444. Each seed uses 500 episodes per LIBERO suite and 500 episodes per Simpler task. We compute mean stall and success rate within each seed, then report the arithmetic mean of the four seed-level values. Each evaluated suite or task thus contributes 2,000 episodes across the four seeds.

\paragraph{Action execution timing.}
\label{sec:action-execution-timing}
We use nominal action durations of 0.1--2.0\,s in simulation, informed by real-world robot-arm systems, including Diffusion Policy's 0.1\,s policy steps on UR5 Push-T~\citep{chi2024diffusionpolicy}, and linear pushing actions executed approximately every 2\,s (0.5\,Hz) on the CloudGripper real-robot platform~\citep{10611548}. MotionServe uses this heterogeneous execution-rate workload to study how robots use execution slack to share GPU inference. Each policy action is decoded once into a target pose under the original controller's action convention. The executor generates a smooth reference trajectory over the configured action duration, which the low-level controller tracks at a fixed frequency while the simulator advances robot dynamics with a fixed physics timestep. Interpolation substeps belong to the same policy action and do not reapply its full displacement increment. We use real-time pacing to synchronize simulated physical time with wall-clock time at a 1:1 rate, so these durations represent physical execution rather than injected artificial delays.

At episode initialization, each robot $i$ samples a nominal action period $\bar\Delta_i$ uniformly from 0.1--2.0\,s and keeps it fixed within that episode.

We introduce execution-time variation following prior work on dynamics randomization~\citep{Sim2Real2018}. The Fetch-arm experiments in that work use $\Delta t=0.04\,\mathrm{s}+\operatorname{Exp}(\lambda)$, with rate $\lambda$ sampled uniformly from $[125,1000]\,\mathrm{s}^{-1}$ at episode initialization and held fixed while the perturbation is resampled at every action step. We retain this rate-selection rule, sampling $\lambda_i\sim\mathcal U(125,1000)\,\mathrm{s}^{-1}$ once per robot per episode. MotionServe uses a robot-specific nominal period and draws one perturbation per action chunk:
\[
 \bar\Delta_i\sim\mathcal U(0.1,2.0)\,\mathrm{s},\qquad
 \epsilon_{i,k}\mid\lambda_i\sim\operatorname{Exp}(\lambda_i),\qquad
 \Delta_{i,k}=\bar\Delta_i+\epsilon_{i,k},
\]
where $k$ indexes chunks and $\operatorname{Exp}$ uses rate parameterization. Each new chunk independently draws its perturbation, and all actions in that chunk share $\Delta_{i,k}$. A full chunk of $H_i$ actions therefore has simulated physical duration $H_i\Delta_{i,k}$. The conditional mean perturbation is $\mathbb E[\epsilon_{i,k}\mid\lambda_i]=1/\lambda_i$, or 1--8\,ms over the stated rate range. These values describe conditional means, not deterministic perturbation bounds. The exponential distribution has unbounded support, so 0.1--2.0\,s bounds only $\bar\Delta_i$, not $\Delta_{i,k}$.

\paragraph{Physical execution and buffer budgets.}
Robots continue executing buffered actions during communication, GPU queueing, and inference, so execution slack originates from actions that remain to be physically executed.

For the current chunk $k$, Equation~4 uses $\Delta_i=\Delta_{i,k}$. The registration-time index $p_i$ gives remaining execution time $(H_i-p_i)\Delta_{i,k}$. After the replacement in Equation~5, the retained $H_i-s$ actions use $\Delta_i^{\mathrm{new}}=\Delta_{i,k+1}$, giving duration $(H_i-s)\Delta_{i,k+1}$. These periods parameterize modeled buffer-depletion deadlines. With the real-time pacing described above, these remaining physical execution intervals provide wall-clock budgets for queueing, inference, and delivery.

To obtain the skip count, the client maintains a completed-action counter $c_i(t)$ and records $c_i^{\rm obs}=c_i(r_i)$ at observation capture. At handoff, $s=c_i(\tau_i)-c_i^{\rm obs}$. The counter advances only when an action completes and pauses during buffer-empty holds, so waiting adds no skipped actions. This count is not inferred from elapsed request latency.

\paragraph{Timing across chunk handoffs.}
Because $\bar\Delta_i$ is fixed within an episode, adjacent chunks differ only through their perturbations:
\[
 \Delta_{i,k+1}-\Delta_{i,k}=\epsilon_{i,k+1}-\epsilon_{i,k}.
\]
For independent, identically distributed exponential perturbations at a common fixed rate $\lambda_i$,
\[
 \mathbb E\!\left[\left|\Delta_{i,k+1}-\Delta_{i,k}\right|\mid\lambda_i\right]
 =\frac{1}{\lambda_i}.
\]
This follows from $|x-y|=x+y-2\min(x,y)$ and the mean $1/(2\lambda_i)$ of the minimum of two independent exponentials. Under these conditions, the expected single-action period difference is 1--8\,ms. For a realized skip count $s$, define the nominal temporal-grid mismatch
\[
 E_{i,k}^{\mathrm{time}}=s\left|\Delta_{i,k+1}-\Delta_{i,k}\right|.
\]
This quantity compares the time assigned to $s$ actions on the two constant-period grids. It captures only period variation, not differences between the old and newly predicted action commands or their resulting states. It is not a bound on trajectory error, and the single-action expectation does not establish negligible mismatch at every handoff.

Equation~5 therefore provides discrete action-step alignment, rather than exact continuous-time trajectory continuity. Overlap-and-skip remains the handoff mechanism: the robot executes the retained suffix from its current physical state. The closed-loop task-success results in Section~4.4 evaluate the full execution loop with this handoff, but do not directly establish trajectory continuity at individual switches.

\subsection{Baseline Adaptation and Component Integration}
\label{sec:baseline-adaptation}

\paragraph{Runtime integration and comparison scope.}
We implement the baseline mechanisms in a shared VLA runtime using the adaptations below. These configurations are used in the baseline experiments and component ablations. For each method, we describe the retained source mechanisms, changes required by chunk-level VLA inference and resident per-robot adapters, and the interface to the client executor or scheduler. Method names refer to these runtime adaptations. Paired ablations hold the surrounding execution and serving configuration fixed while changing the component under study.

\paragraph{Common runtime boundary.}
Each configuration uses the same model checkpoint, per-robot adapter assignment, observation preprocessing, action decoding, task allocation, and execution-time distribution. A request carries a robot ID, adapter ID, generation ID, observation, capture-time action counter, and buffer metadata. The model runner returns a complete action chunk and any method-specific horizon metadata. The client executor determines when to observe and which actions remain executable. The scheduler determines request admission, batch membership, and dispatch time. Model weights, numerical precision, and decoding settings are held fixed within each paired comparison. All configurations have the same outer limit of $N$ robot requests per batch. A method may choose smaller batches. Internal samples, stage workspaces, and concurrent stages still count toward actual GPU resource use.

For scheduling comparisons, execution metadata and the same calibration inputs are available to every scheduler that needs them. Let $D_i$ be the depletion deadline of the actions the client actually permits to execute, and let $G_i=D_i-\delta_i$ be the corresponding GPU-completion deadline. Elapsed upload time is included when converting the remaining budget to the server clock. Deadlines use no future inference times, future execution perturbations, or evaluation outcomes. Method-specific profiling includes the actual runner and horizon-selection work, including AAC's multiple samples and Robion's concurrent stages.

\paragraph{Conventional schedulers.}
The FCFS port is work-conserving: whenever the runner becomes available, it takes the oldest compatible pending requests up to the cap, without an intentional accumulation delay. Fixed Window uses the same arrival order with a fixed maximum batching delay $W$, measured from the oldest waiting request. It commits when the cap is reached or that delay expires, and never restarts the timer on a new arrival. This implements a fixed-window variant of the delayed batching discussed in Clipper~\citep{201468} within the shared runner. EDF orders requests by increasing $G_i$, breaks ties by arrival order, and dispatches when the runner is available. It does not use SlackBatch's grouping or slack-dependent waiting. The fixed-window value and any preferred batch sizes are selected on calibration workloads, frozen before evaluation, and kept identical in each with/without-MotionExec pair.

\paragraph{Armory: a Lookahead scheduling port.}
Armory maintains a mirror of robot action queues and evaluates future batches by predicted robot execution progress~\citep{bansal2026actionchunkschedulingbatched}. Our port retains its candidate-batch search, one-step lookahead ($L=1$), equal robot weights ($w_i=1$), and the source implementation's one-second execution-scoring window and GPU-time normalization. The mirror receives completed-action counters, retained chunk ranges, and measured communication and inference costs. It uses each robot's action duration and simulates the common client's handoff and starvation behavior. CPU search proceeds while the GPU is busy and dispatches the best available useful batch when it becomes idle. The common client determines request eligibility. Native Armory streams images at each step and replaces stale observations with the latest image. Our port instead serves the observation attached to the eligible request. This configuration evaluates Armory's Lookahead scheduler under the common observation and handoff mechanism.

\paragraph{Robion: stage execution and SLO scheduling.}
Robion's central mechanism combines two GPU streams, fixed batch membership within each stage, lockstep execution, and profiled SM partitioning~\citep{adamopoulos2026efficientvisionlanguageactionmanagementserving}. Our port retains these mechanisms: the vision--language stage of one batch overlaps the action stage of the preceding batch, with the next lockstep starting after both complete. X-VLA uses its flow-matching action stage. OpenVLA-OFT uses its parallel regression head~\citep{KimM1-RSS-25} as the action stage in place of an ADiT. Stage boundaries preserve each model's native intermediate tensors and outputs, and we remeasure the stage-pair/batch-size profiles on H200. The port keeps per-adapter FIFO queues, prioritizes their heads by remaining $G_i-t$, and admits compatible heads into a mixed-adapter stage batch. This explicitly extends the source's single-model-per-batch rule to the shared-backbone runner. Single-GPU placement is fixed, so multi-GPU traffic assignment is inactive.

\paragraph{CoLoRA: coordinated priority and batch formation.}
CoLoRA coordinates adaptive request priority (APS), adapter-aware placement/cache management (AAS), and load-aware batching (LBS)~\citep{lin2026colora}. In our port, queue age maps to request waiting time, SLA urgency to the remaining GPU-completion budget, and adapter identity to the robot's policy ID. The unified scheduler refreshes these signals and GPU-load/queue-depth measurements before applying APS and LBS. All adapters are resident on one GPU, so residency preferences and eviction decisions become inactive. We retain adapter grouping within a batch but allow several adapter groups to share one invocation. This is an explicit mixed-adapter extension. Restricting each batch to one adapter would yield singletons when every robot has a unique adapter and at most one outstanding request. The comparison therefore evaluates coordinated priority and load-aware batch formation with resident policies in the shared runner.

\paragraph{S-LoRA: chunk-level multi-adapter admission.}
S-LoRA combines heterogeneous-adapter batching, iteration-level admission, and paged memory management. It also studies adapter clustering and early abort~\citep{MLSYS2024_906419cd}. Our adaptation retains heterogeneous-adapter batching but performs admission only at complete action-chunk boundaries, using the common mixed-adapter runner. It does not treat successive flow steps as autoregressive token iterations. Because adapters are resident, adapter swapping is inactive. We disable early abort for closed-loop execution: an overdue robot request remains serviceable, with its continued waiting counted as stall.

\paragraph{Chameleon: resource-aware multi-queue admission.}
Chameleon classifies requests by estimated work and admits them using per-queue quotas followed by redistribution of unused capacity~\citep{Iliakopoulou_2025}. Our port retains this two-pass admission procedure and non-preemptive execution. Its language-model output-length predictor is replaced by known action-generation work: the model's fixed prediction horizon, sampling/denoising configuration, input shape, and adapter rank. Calibration maps these features to estimated execution cost for queue assignment and measured transient-memory requirements for quota accounting. Queue cutoffs and quota rules are configured on the calibration workload and applied consistently across runs. All adapters are resident, so adapter-cache eviction and cache-miss bypassing are inactive. Queue assignment uses computational cost, with equal-cost requests occupying the same queue. This configuration evaluates Chameleon's admission mechanism in the resident-policy setting.

\paragraph{A common pipelined wrapper for horizon selectors.}
PACE, AAC, and AutoHorizon choose how much of a predicted chunk to execute before replanning~\citep{nie2026pace,Liang_2026_CVPR,10.1007/978-3-032-37196-6_33}. For the scheduler ablations, we wrap each selector $g$ in the same pipelined action executor $E(g)$. It interprets the returned action count $h_i=g(A_i)\in[1,H_i]$ as an observation trigger. After $h_i$ actions have completed, the client captures an observation while retaining the old suffix for execution during service. It allows one outstanding request and applies the completed-action skip and atomic buffer replacement in Equation~5. If the trigger has already been skipped at installation, it requests at the next eligible boundary. If no returned action remains, it holds and requests again without treating waiting as completed execution.

This wrapper converts an execution-horizon selector into a request-trigger selector. Native prefix execution discards the suffix at $h_i$. $E(g)$ can execute beyond $h_i$ while waiting. Both members of each with/without-SlackBatch pair use $E(g)$. Its remaining execution budget is $(H_i-p_i)\Delta_i$ because the suffix remains executable. Under native prefix execution, the budget would instead be $\max(0,h_i-p_i)\Delta_i$, reaching zero when a request is issued after that prefix. The reported pairs use the retained-suffix budget consistently in both configurations.

\paragraph{PACE trigger port.}
PACE detects prominent low-speed valleys and selects the earliest accepted boundary, falling back to its maximum horizon when none is accepted~\citep{nie2026pace}. For our single-arm Cartesian policies, our port constructs the speed profile from decoded end-effector positions, reconstructing positions first when the action interface supplies increments. Gripper channels are excluded from the speed norm. For X-VLA's 30-action chunks, we use the released implementation's 11-sample Savitzky--Golay window, polynomial order 3, and minimum valley separation of 10, with maximum horizon 30. Task thresholds use the fifth percentile of positive demonstration-segment contrasts, following the released calibration procedure, with demonstration and prediction profiles expressed at the same reference sampling rate. These settings are frozen before evaluation. The returned horizon is an action count, not a zero-based trigger index. Both PACE rows use $E(g_{\rm PACE})$. Neither uses MotionExec's two-segment objective.

\paragraph{AAC trigger port.}
AAC estimates translation/rotation uncertainty and gripper entropy from multiple action samples, then chooses the prefix at the largest increase in cumulative mean entropy, subject to its minimum-motion constraint~\citep{Liang_2026_CVPR}. Our X-VLA port uses $m=20$ independently sampled chunks per observation, matching the paper's reference sample count. It uses the released Gaussian/Bernoulli estimator, equal component weights, $10^{-6}I$ covariance regularization in normalized coordinates, and candidate 0 for execution in both rows. Action channels are converted to the estimator's translation, rotation, and binary-gripper conventions. The minimum-motion calculation uses decoded physical displacements rather than sums of absolute poses, with its threshold calibrated in those units. All horizon searches use the actual $H_i=30$, replacing source-client constants for 16-action policies. Entropy is estimated across samples of one robot, never across robots. A batch of $b$ requests contains $mb$ action samples, even if observation features can be reused. Sampling, entropy calculation, transfers, and microbatching are included in measured service cost and the AAC-specific SlackBatch profile. The cap counts robots, not samples.

\paragraph{AutoHorizon trigger port.}
AutoHorizon derives an execution horizon from action self-attention using entropy filtering and bidirectional soft pointers~\citep{10.1007/978-3-032-37196-6_33}. Our X-VLA port extracts the action-to-action attention block at the third flow-sampling step, averages over layers and heads separately for each robot, and applies row normalization before horizon selection. It uses the source defaults $q=0.9$ and $\tau=0.3$. Image, language, state, and soft-prompt positions are excluded from the action block, and the robot batch axis is never averaged away. Any attention-kernel change, materialization, reduction, and transfer cost is included in both rows and their service profile. The port preserves the original action predictions, prediction horizon, and trained weights while supplying the horizon selector to the shared wrapper.

\paragraph{Adding MotionExec and SlackBatch.}
For each conventional scheduler $S\in\{\mathrm{FCFS},\mathrm{Fixed\ Window},\mathrm{EDF}\}$, the matched control uses a fixed trigger $h_0$ inside the same pipelined wrapper, with $h_0$ selected on calibration episodes and frozen across loads. The pair is $(E(h_0),S)$ versus $(E(g_{\rm ME}),S)$, where $g_{\rm ME}$ returns the action count corresponding to MotionExec's sweet-spot boundary. Only the motion-based trigger changes. The scheduler implementation, batching timer, cap, handoff, and overdue-request handling remain fixed. This isolates the benefit of motion-guided observation timing within the pipelined action executor.

For each $g\in\{g_{\rm PACE},g_{\rm AAC},g_{\rm AutoHorizon}\}$, the scheduler pair is $(E(g),\mathrm{FCFS})$ versus $(E(g),\mathrm{SlackBatch})$. The selector, client wrapper, sample count, action head, and profiling procedure are identical within the pair. SlackBatch replaces only admission, waiting, and batch selection, and obtains budgets from that executor's actual remaining actions. MotionServe is $(E(g_{\rm ME}),\mathrm{SlackBatch})$. For isolated comparisons of Armory-Lookahead and the multi-LoRA scheduling ports against SlackBatch, $E(g_{\rm ME})$ is likewise held fixed. Robion's comparison includes its two-stage execution engine as well as its request-selection mechanism.

MotionExec retains $\sigma_p=0.01\,\mathrm{m}$, $\sigma_R=0.10\,\mathrm{rad}$, and gripper weight 4 in Equation~2, with earliest-index tie-breaking. SlackBatch reevaluates on request arrival, GPU completion, and waiting expiry, with a nominal 5\,ms expiry poll. Its committed batches execute through FIFO $Q_C$, with delivery overlapping subsequent inference. Initialization is separate: the recorded counts are 1, 1, 1, 2, and 3 batches at the five fleet sizes and are excluded from batch-size summaries.

\paragraph{Paired experimental controls.}
Paired runs use matching task states and random seeds for policy sampling and action timing, the same episode budgets, and the same resource limits. Calibration settings are fixed before evaluation. Within each pair, method-specific parameters, the client wrapper, and the profiling procedure remain unchanged except for the component identified above. Stall, retries, initialization, and auxiliary CPU/GPU work use the same accounting conventions. CoLoRA and Robion use the mixed-adapter extensions described above in all reported configurations. These controls isolate the stated component changes within our runtime. The scope of each source-system comparison is determined by its retained mechanisms and explicit adaptations.

\subsection{Offline GPU Inference Profiling}

Before online serving begins, we run each model on the inference server at every integer batch size $b=1,\ldots,N$ and measure its GPU inference time, where $N$ is the number of robots and the batch-size cap. These measurements are completed in advance and used to construct SlackBatch's fixed inference-time lookup table $\widehat\ell(b)$. During online serving, the scheduler queries this table by batch size to estimate GPU work.

The same offline trials separately measure each robot's result-delivery latency, from GPU completion to receipt of the complete action chunk and sweet-spot metadata. The resulting fixed estimate $\delta_i$ is deducted from robot $i$'s remaining execution budget to reserve time for result delivery. SlackBatch uses the remaining budget for GPU waiting and inference.

\newpage
\section{Distribution of Robot Stall Times}
\begin{figure}[!ht]
 \centering
 \includegraphics[width=\textwidth]{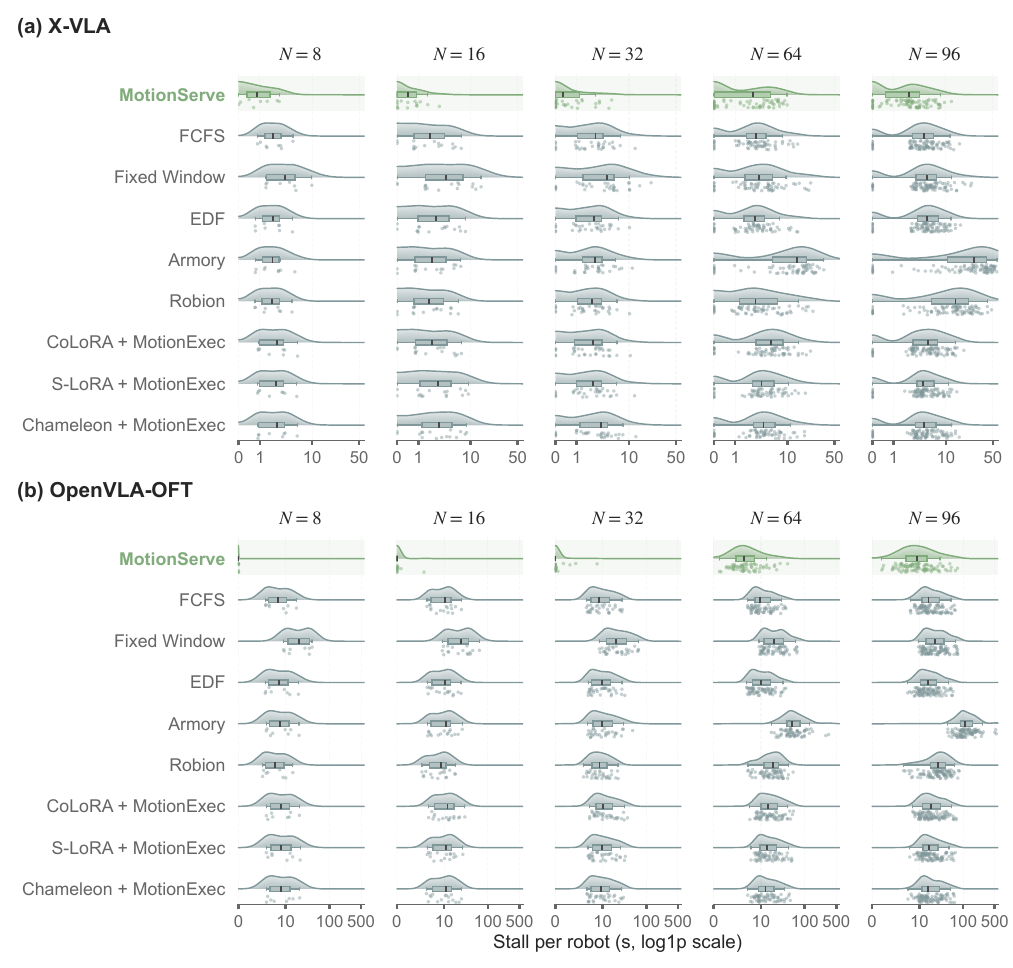}
 \caption{\textbf{Robot stall distributions.} The panels compare X-VLA and OpenVLA-OFT across robot counts, with MotionServe highlighted in green. Each point represents a robot's accumulated mean stall time per episode, while box plots summarize the median and interquartile range and half-violins show the density. Stall times are plotted on a $\log(1+\mathrm{stall}/1\,\mathrm{s})$ scale with tick labels in seconds.}
 \label{fig:robot-stall-distribution}
\end{figure}

Figure~\ref{fig:robot-stall-distribution} complements the fleet-wide means in Section~4.3 by showing the per-episode stall of robots with heterogeneous action periods at each model--method--load condition. MotionServe has the lowest median stall at every fleet size for both models. At $N=96$, its median stall is 2.30\,s for X-VLA and 8.75\,s for OpenVLA-OFT, compared with the lowest baseline medians of 4.21\,s for S-LoRA + MotionExec and 15.92\,s for Chameleon + MotionExec. The distributions expose variation within a fleet.
\label{sec:complete-results}
